\documentclass{amsart}

\usepackage{amssymb,latexsym,amsfonts,amsmath}
\usepackage{graphicx}
\usepackage{eso-pic}
\usepackage{epsfig}
\usepackage{graphicx}
\usepackage{xcolor}
\usepackage{algorithm}
\usepackage{algorithmic}

\usepackage{amssymb,latexsym,amsfonts,amsmath}
\usepackage{graphicx}
\usepackage{subcaption}
\usepackage{amsthm}
\usepackage{amsmath}

\newtheorem{theorem}{Theorem}[section]

\newtheorem{problem}[theorem]{Problem}

\theoremstyle{definition}
\newtheorem{definition}[theorem]{Definition}
\newtheorem{example}[theorem]{Example}

\theoremstyle{remark}
\newtheorem{remark}[theorem]{Remark}
\numberwithin{equation}{section}

\begin{document}

\begin{abstract}
Reward machines (RMs) provide a structured way to specify non‑Markovian rewards in reinforcement learning (RL), thereby improving both expressiveness and programmability. Viewed more broadly, they separate what is known about the environment, captured by the reward mechanism, from what remains unknown and must be discovered through sampling. 
This separation supports techniques such as \emph{counterfactual experience} generation and \emph{reward shaping}, which reduce sample complexity and speed up learning.

We introduce physics-informed reward machines (pRMs), a symbolic machine designed to express complex learning objectives and reward structures for RL agents, thereby enabling more programmable, expressive, and efficient learning.
We present RL algorithms capable of exploiting pRMs via counterfactual experiences and reward shaping. Our experimental results show that these techniques accelerate reward acquisition during the training phases of RL.
We demonstrate the expressiveness and effectiveness of pRMs through experiments in both finite and continuous physical environments, illustrating that incorporating pRMs significantly improves learning efficiency across several control tasks.
\end{abstract}


\title[Physics-Informed Reward Machines]{Physics-Informed Reward Machines}

\author[daniel ajeleye]{Daniel Ajeleye} 
\author[ashutosh trivedi]{Ashutosh Trivedi}
\author[majid zamani]{Majid Zamani} 
\address{Department of Computer Science, University of Colorado Boulder, USA}
\email{\{daniel.ajeleye, ashutosh.trivedi, majid.zamani\}@colorado.edu}

\maketitle

\section{Introduction}
\label{sec1}
Reinforcement learning (RL)~\cite{sutton2018reinforcement} is a learning-based optimization approach in which an agent repeatedly interacts with its environment to converge on optimal strategies, guided by scalar reward signals.
A key appeal of the RL paradigm is that the learning agent does not require a model of the environment---the transition probabilities and reward mechanism---and instead only needs access to a mechanism that can sample experiences from the environment.
While it is reasonable to assume that transition probabilities are unknown, there is less justification for concealing the reward mechanism from the agent, especially when it is complex, structured, or based on logical formalism.
It is around this distinction that reward machines (RM)~\cite{toro2019learning} have been proposed to expose the learning objective to the agent in the form of a finite-state transition mechanism.

The key semantic role of RMs, apart from allowing designers to express non-Markovian learning objectives, lies in providing the learning agent with a high-fidelity model of some aspects of the environment.
From this perspective, we posit that reward machines provide a convenient interface to reveal other high-fidelity information about the physical aspects (\emph{e.g.,} differential equations, conservation laws, heat and chemical reaction laws) of the environment to the agent, thereby improving sampling efficiency and explainability by influencing strategy representation.
We refer to these as \emph{physics-informed reward machines (pRMs)}.

\paragraph{Physics-Informed Neural Networks.} Significant improvements in the performance of supervised learning have been achieved through the incorporation of additional physical information, as demonstrated in Physics-Informed Neural Networks (PINNs)\cite{cai2021physics,cuomo2022scientific,pang2019fpinns,krishnapriyan2021characterizing,raissi2019physics}. 
This information, typically encoded within the loss functions during training, aims to expose well-known aspects of physics to the network, thereby reducing the strain on the learning model.
PINNs have also been gaining traction~\cite{chen2022physics,chentanez2018physics,cho2019physics,gokhale2022physq,jiang2022physics,mukherjee2023bridging,ramesh2023physics} within RL, similarly leveraging physics in the context of deep RL---appropriately termed physics-informed RL (PIRL) (for a recent survey, see~\cite{banerjee2023survey})---to improve sample complexity, training speed, and explainability.

Our goal differs from the aforementioned literature on PINN and PIRL. 
While our approach can be combined with physics-informed training of various neural networks in deep RL, our primary focus is on providing a partial high-fidelity model to the RL agent, with the aim of achieving both speedup and enhanced explainability. 
PINNs primarily focus on approximating solutions to physical equations using neural networks, while PIRL embeds physics directly into the RL process to guide agent learning. In contrast, this work centers on pRMs, which integrate physical dynamics directly into the reward structure for RL. Unlike PINNs, pRMs do not model the state evolution of the environment. Instead, they provide a structured and high-fidelity representation of the reward function. By encoding physics into both the reward structure and logical task specifications, pRMs enhance task specification and learning efficiency. This added expressiveness enables encoding complex tasks with logical structures, surpassing the implicit task logic in reward functions or the procedural definitions typical of PIRL approaches. Our approach bridges formal RL reward structure representations and physical dynamics, promoting interpretable, sample-efficient, and explainable learning.

\paragraph{Physics-Informed Reward Machines (pRMs).}
Reward machines (RMs)~\cite{icarte2022reward,toro2019learning} were proposed to express non-Markovian aspects of rewards in RL.
In formal methods and control design communities, the reward structures are often expressed in formal logic or $\omega$-automata based formalisms ~\cite{sadigh2014learning,hahn2019omega,alur2023policy,camacho2019ltl}, which can then be automatically compiled into finite-state RMs \cite{middleton2020fl}. 
We adopt the RM perspective to expose physical modeling to the RL agent, leveraging it to generate additional counterfactual experiments based on physical dynamics.
We introduce a symbolic form of RM that we call physics-informed reward machines (pRM). 

\begin{figure}[t!]
    \centering
    \caption{An Office-World Example.}
    \label{fig11}  
    \includegraphics[width=0.5\linewidth]{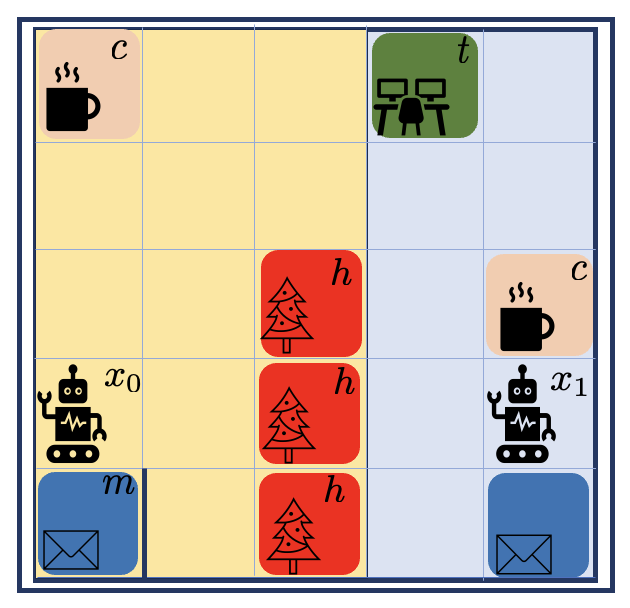}
\end{figure}

These pRMs offer a structured framework for specifying complex reward functions and control tasks for RL agents.
They support flexible composition of reward specifications, including concatenation, loops, conditional logic, and enforcement of regular and physically motivated constraints.
As the agent interacts with a stochastic environment, it progresses through the states of the pRM, with transitions triggered by observed events and constrained by task specifications.
At each transition, the pRM supplies the relevant reward function for the agent to use.
This approach integrates known physical structure with real-time task execution, enhancing the effectiveness of RL in physical domains.

\begin{example}[Office World Example]
    \label{mot_ex}
As a motivating example for pRMs, consider the Office World environment. The RL agent operates in a grid-based setting, illustrated in Figure~\ref{fig11}, where each cell corresponds to a state. Specific states are labeled $x_0$, $c$, $m$, $h$, and $t$, denoting the initial state, coffee pickup, mail pickup, ornament location, and the delivery destination, respectively. These labels represent events triggered when the agent visits the corresponding states. At each state, the agent can select one of four cardinal-direction actions. In the $h$ and $t$ states, all actions leave the agent stationary. In all other states, the agent moves in the chosen direction with probability $\frac{1}{3}$, and in each perpendicular direction with probability $\frac{1}{3}$. If a movement would take the agent into a wall, it remains in its current state.

Let the yellow and blue regions in the environment represent warm and cold areas, respectively. Suppose the agent's task is to collect both the mail and the coffee and deliver them to the office without colliding with any decorations. Such tasks can be naturally specified using traditional RMs. 

However, in this work, we focus on a more nuanced task where the agent has the knowledge about the physical dynamics affecting the coffee’s temperature---such as heating or cooling behavior determined by the cup's material and the region of the environment in which the agent is located. Specifically, the coffee's temperature evolves based on whether the agent is in a warm or cold region. In this setting, we aim for the reward machine to leverage this known physical information---namely, the temperature dynamics of the coffee cup based on the agent's location---to guide the agent toward delivering the coffee at a desirable temperature, despite the underlying environment dynamics being unknown. 
As illustrated in Figure~\ref{mot_ex}, if the agent starts from position $x_0$, it can successfully deliver the coffee within eight steps at the desired temperature, informed by its awareness of being in a warm region (as detected by the cup's sensor). In contrast, starting from point $x_1$, even a six-step plan may lead to an undesirable delivery due to unaccounted cooling effects. 
\end{example}

This example illustrates how temporal and physical constraints can be jointly captured using pRMs for task specification and reward assignment in unknown environments. Inclusion of physical dynamics introduces additional complexity and noise into the learning process---scenarios that traditional RMs are not equipped to handle. While standard RMs enable agents to learn policies that satisfy the logical sequencing of tasks, pRMs go further by incorporating physical properties, such as whether the coffee remains at an acceptable temperature upon delivery. The agent's ability to reason about such physical information not only enhances sample efficiency but also yields more interpretable and desired behaviors. The development of methods for effectively exploiting this enhanced expressiveness of pRMs forms the central focus of this work.

\paragraph{Contributions.} Our key contributions are as follows.
\begin{itemize}
    \item[1.] We introduce physics-informed Reward Machines (pRMs) that incorporates known physical dynamics into the reward structure in RL. We demonstrate how the expressive structure of pRMs enables the integration of physical laws with real-time dynamics in reinforcement learning tasks, thereby improving sample efficiency and enhancing the interpretability of learned policies.
    \item[2.] We extend existing techniques for exploiting traditional reward machines to the pRM setting, including the use of counterfactual experience generation guided by physical dynamics and the application of reward shaping to address reward sparsity. We also discuss convergence conditions under which these methods remain effective.
    \item[3.] We empirically evaluate the proposed framework across both finite and continuous physical environments, including the office gridworld, two-tank system, five-room temperature model, and five-road traffic network. Results show that pRMs significantly improve sample efficiency and contribute to the interpretability of policies in complex stochastic control tasks.
\end{itemize}

\section{Preliminaries}
Given $M$ vectors $v_i\in \mathbb{R}^{m_i}$, $m_i\in \mathbb{N}$, and $i\in[1;N]$, $v = [v_1;\cdots; v_M]$ indicates the corresponding column vector of dimension $\sum_i m_i$. 
For $p \geq 1$ and $w \in S$, $w^{\le p}$ represents a sequence of $w$'s of length at most $p$, while $S^\omega$ denotes the set of $\omega$-sequences from $S$. For any vectors $a$ and $b$ in the same vector space and $\kappa \in [0,1]$, the notation $a \stackrel{\kappa}{\leftarrow} b$ denotes $a \leftarrow a + \kappa \cdot (b - a)$ (used in Alg. \ref{alg1} and \ref{alg2}).

A discrete \emph{probability distribution} over a set $S$ is a function
$d \colon S {\to} [0, 1]$ such that $\sum_{s \in S} d(s) = 1$ and $\textrm{supp}(d) = \{s \in S\::\: d(s) > 0\}$ is at most countable.  Let
$\mathcal{D}(S)$ denote the set of all discrete distributions over $S$.  

\begin{definition}[Markov decision process]
    \label{mdp}
    A Markov decision process (MDP) is a tuple $\mathcal{M} = (X,U,r,\mathbb{P},\lambda)$ where: 
    \begin{itemize} 
    \item $X$ is an (uncountably infinite) set of states,
    \item $U$ is an (uncountably infinite) set of actions, 
    \item $\mathbb{P}: X {\times} U \to \mathcal{D}(X)$ is the  transition probabilities, 
    \item $r:X\times U\times X\rightarrow \mathbb{R}$ is the reward function, and
    \item $\lambda\in(0,1]$ is the discount factor.
    \end{itemize}
\end{definition}

\noindent 
For $(x, u, x')\in X {\times} U {\times} X$, we write $\mathbb{P}[x' | x, u]$ as a shorthand for $\mathbb{P}(x, u)(x')$.
We say that an MDP is \emph{finite} if both $\lvert X\rvert$ and $\lvert U\rvert$ are finite.
When the transition probability $\mathbb{P}$  is not explicitly available and can only be accessed through sampling, we refer to the MDP as \emph{unknown}.

\paragraph{Reinforcement Learning (RL).} RL provides a sampling-based approach to optimization where a \emph{learning agent} samples the unknown MDP over several episodic interactions to compute policies maximizing discounted rewards.  
Formally, a Markov policy $\rho:=\langle \rho_k:X{\rightarrow} U\rangle_{k\geq0}$ for an MDP $\mathcal{M}$ is defined as a probability distribution over actions in $U$ given a state $x \in X$. At time step $k$, the current state of the agent is $x_k \in X$. If the agent selects an action $u_k \in U$ according to $\rho_k(\cdot \mid x_k)$ and executes $u_k$, it transitions to a new state $x_{k+1}$ according to the transition probability $\mathbb{P}[\cdot \mid x_k, u_k]$ and receives a reward $r_k := r(x_k, u_k, x_{k+1})$ from the environment. In a \emph{training episode}, this process is repeated starting from $x_{k+1}$ and continues until certain termination conditions are satisfied, such as reaching a maximum number of time steps or a sample threshold.
From any initial state $x_0$, the agent aims to learn an optimal policy $\rho^*$ that maximizes the expected discounted return:
\[
R := \mathbb{E}_{\rho^*} \left[\sum_{i=0}^\infty \lambda^i r_{i} \big| x_0\right].
\]
Although our goal is not to construct (finite) abstractions of MDPs, it is worth noting that infinite MDPs can be approximated (in a model-free sense) by finite ones, provided that discretization of the state-action space $(X \times U)$ is computationally feasible. Several methods for constructing such abstractions are available in the literature; see, for instance,~\cite{li2006towards,lavaei2020formal,ajeleye2023data,ajeleye2024bdata}. 
Dealing with a finite MDP is often valuable because convergent model-free tabular RL algorithms, such as Q-Learning (QL) \cite{watkins1992q,borkar2000ode}, are well-suited to finite MDPs. 

QL is inapplicable for problems involving infinite MDPs. In such settings, function approximation methods become essential. A case in point is the Deep Deterministic Policy Gradient (DDPG) algorithm \cite{lillicrap2015continuous}, which simultaneously learns both the $Q$-function and the policy. 

\paragraph{Reward Machines.} Complex learning objectives often cannot be captured using Markovian reward signals as defined in standard MDPs. A recent approach to address this limitation is the use of finite-state reward machines (RMs), which provide a structured way to specify non-Markovian rewards~\cite{icarte2022reward}.

A reward machine is a tuple $\mathcal{R} = (\Delta, \mathcal{Q}, \varrho_0, \rho_{\varrho}, \rho_r)$, where $\Delta$ is a set of atomic propositions, $\mathcal{Q}$ is a finite set of states, $\varrho_0 {\in} \mathcal{Q}$ is the initial state,
$\rho_\varrho\colon \mathcal{Q} {\times} 2^\Delta \to \mathcal{Q}$ defines the transition relation, 
and $\rho_r\colon \mathcal{Q} \times 2^\Delta \times \mathcal{Q} \to \mathbb{R}$ is the reward function.
Due to their finite-state nature, RMs can be composed with finite MDPs to yield an equivalent MDP with Markovian rewards. However, this composition need not be performed explicitly; instead, an RL agent can use the RM as an online monitor to generate non-Markovian rewards. The explicit structure of RMs also allows for counterfactual queries~\cite{icarte2018using}, improving sample efficiency.

As discussed in the introduction, integrating continuous dynamics into RMs enables the specification of continuously evolving reward logic. Furthermore, RMs facilitate the transfer of high-level task specifications from designers to RL agents. By extending RMs with physical information, this framework can incorporate known dynamics into the reward specification. In the next section, we introduce physics-informed reward machines and present RL algorithms designed to leverage the physical structure embedded in them.

\section{Physics-Informed Reward Machines}
\label{sec_rm}
We propose \emph{physics-informed reward machines (pRMs)}, a novel RM framework that enhances expressiveness by incorporating physical knowledge into the reward structure. This hybrid configuration enables agents to interact more effectively with their environment, facilitating faster reward acquisition and policy learning.

Let $\Delta$ be a set of atomic propositions. We assume that there are labels of events in $2^\Delta$ that are of high-level of relevance in a region of the environment state set, and the agent can perfectly detect those labels. Therefore, we introduce a labeling function over the state set $X$, namely $L:X\rightarrow 2^\Delta$. The labeling function is useful for providing truth assignments to a state successor under an action, and such a transition of the environment can thereby be input to the pRM in order to reward the agent. Hence, the pRM is defined over the set of propositions in $2^\Delta$. Next we formally define pRMs to handle reward function structures in this work.
\begin{definition}[Physics-Informed Reward Machines]
\label{rm}
Given an unknown MDP $\mathcal{M}$, 
let $\Delta$ and $L:X\rightarrow 2^\Delta$ be a set of propositional symbols and a labeling function, respectively. A physics-informed reward machine (pRM) $\mathcal{A}_R$ is a tuple $(\Omega,\Omega_F,\Omega_0,\delta_\varrho, \delta_r)$, where: 
\begin{itemize}
    \item $\Omega\subseteq\{\varrho_i~|~i\in[0;M]\}\times\mathbb{R}^{\ell}$ is the set of states of the pRM and $M,\ell\in\mathbb{N}_{\ge1}$, 
    \item $\Omega_F\subset\Omega$ is the set of \emph{terminal} states, 
    \item $\Omega_0\subseteq(\Omega\setminus\Omega_F)$ is the set of initial states, 
    \item $\delta_r:\Omega\times2^\Delta\rightarrow \mathbb{R}$ is the pRM state-reward function, and
    \item $\delta_\varrho:\Omega\times2^\Delta\rightarrow\Omega$ is the pRM state-transition function.
    \end{itemize}
\end{definition}
Consider an observed experience $(x_k,u_k,x_{k+1})\in X\times U\times X$ of the environment $\mathcal{M}$ at time-step $k\in\mathbb{N}$, where $\mathbb{P}[x_{k+1}~|~x_k
, u_k]>0$. The pRM $\mathcal{A}_R$ receives a propositional symbol $\varphi_k=L(x_{k+1})\in2^\Delta$ based on labeling function $L$ and evolves to state 
\begin{equation}
    \label{prm_dyn}
    \tilde\varrho(k+1)=\delta_\varrho(\tilde\varrho(k),\varphi_k),
\end{equation}
rewarding the agent according to the function 
\begin{equation}
    \label{prm_rew}
r_k=\delta_r(\tilde\varrho(k),\varphi_k).
\end{equation}
Let $g_i:\mathbb{R}^\ell\rightarrow\mathbb{R}^\ell$, $i\in[0;M]$, be absolutely continuous functions. At time-step $k\in\mathbb{N}$, the pRM's current state $\tilde\varrho(k):=(\varrho_i,\psi( k))\in\Omega$ has an hybrid configuration, where $\psi(k)\in\mathbb{R}^\ell$, is a solution of the following ordinary differential equation (ODE):
\begin{equation}
    \label{prm_ode}
\mathrm{d}\psi(k)=g_i(\psi(k))\,\mathrm{d}k.
    \end{equation}

Note that the initial condition of the ODE in \ref{prm_ode} is determined by the starting value of the variable $\psi$, and it reflects the physical state of the system at the starting time and is governed by the particular physical law or process the ODE is modeling. The evolution of the pRM continues as described until it reaches a state in  $\Omega_F$. The terminal states are blocking states where the agent's path on the pRM can be trapped indefinitely. It is important to note that by defining set $\Omega_F$ as empty, the pRM can manage tasks that continue indefinitely without trapping the agent in any particular state. A pRM can be considered as an MDP with a state set $\Omega$ and an action set $2^\Delta$, evolving according to \eqref{prm_dyn}. 

It is important to note that, given a pRM, a practical approach is to discretize its continuous components (\emph{i.e.,} $\Omega$) and to uniformly sample experiences generated by the resulting discrete machine---an approach adopted in this work. While alternative methods may exist for utilizing pRMs without discretization, their investigation lies beyond the scope of this work and is therefore not pursued here.

\begin{figure}[t!]
        \centering
\includegraphics[width=0.75\linewidth]{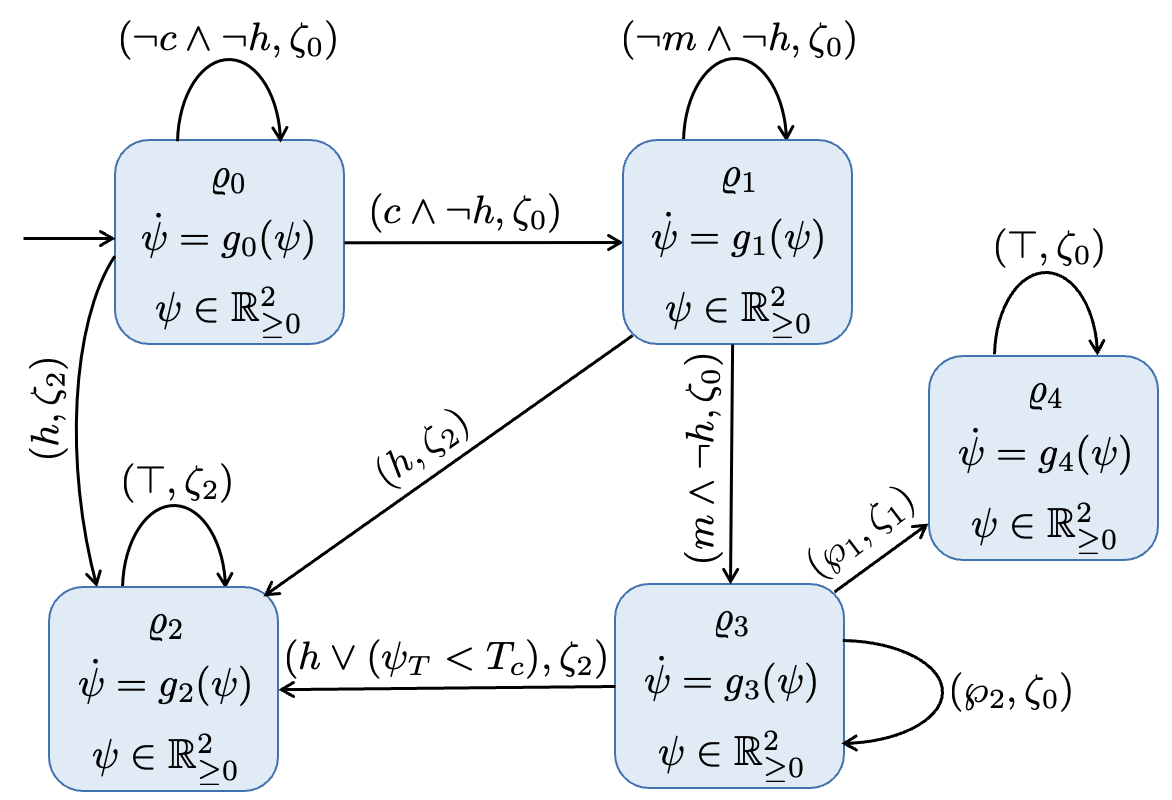}
    \caption{pRM $\mathcal{A}_{R2}$ where $\psi=[\psi_c;\psi_T]$, $\wp_1=t\land\neg h\land (\psi_T\in[T_c,T_0])$ and $\wp_2=\neg t\land\neg h\land (\psi_T\in[T_c,T_0])$.}
        \label{fig_rm2}
\end{figure}

\begin{example}[Motivating Example]
    The pRM $\mathcal{A}_{R2}$, shown in Figure~\ref{fig_rm2}, specifies a variant of the task described in Example~\ref{mot_ex}. Here, the set of atomic propositions is $\Delta=\{c,m,h,t\}$, while the set of states is $\Omega=\{\varrho_i~|~i\in[0;4]\}\times\mathbb{R}^2_{\ge0}$. The terminal states set $\Omega_F=\{(\varrho_2,(\psi_c,\psi_T))~|~\psi_c\in[0,N_c]\text{ and }\psi_T<T_0\}$ and the initial state set $\Omega_0=\{(\varrho_0,(0,T_0))\}$. The corresponding ODEs according to \eqref{prm_ode} are as follows:
    \begin{align}
        \label{rm2_ode}
        g_0(\psi)=\begin{bmatrix}
 1&T_0
\end{bmatrix}^\top\text{, }g_2(\psi)=\begin{bmatrix}
 1&0
\end{bmatrix}^\top,\notag \\
g_i(\psi(k))=\begin{bmatrix}
 1&-\alpha (\psi_T(k)-T_e)
\end{bmatrix}^\top~\forall i\in\{1,3,4\},
    \end{align}
   where $\psi(k)=[\psi_c(k);\psi_T(k)]$, $\psi_c$ is the time counter variable for the label $c$, $\alpha$ is the heat exchange coefficient, $T_e$ is the ambient temperature, $T_0$ is the initial temperature of the coffee, and $\psi_T(k)$ is the coffee temperature at time $k$. The ODE in \eqref{rm2_ode} is sampled at time-step intervals $\tau_{\mathcal{A}}=1s$. Each edge of the graph in Figure~\ref{fig_rm2} representing $\mathcal{A}_{R2}$ has a pair consisting of the observed proposition from the environment using the labeling function, and the assigned reward. Therefore, pRM $\mathcal{A}_{R2}$ specifies the following task: deliver coffee and mail without colliding with flowers, while ensuring that the coffee's temperature remains above $T_c$$^\circ C$.
   Additional illustrative examples of pRMs are provided in the Appendix.
\end{example}

\begin{problem}[RL with pRMs]
    \label{p1}
    Let $\mathcal{M}$ be an unknown MDP, and $\mathcal{A}_{R}$ be a physics-informed reward machine. 
    We are interested in enabling off-the-shelf RL algorithms to compute approximately optimal Markov policy $\rho$ over  $\mathcal{M}$ against reward from $\mathcal{A}_{R}$ in a sample-efficient manner. 
\end{problem}

\paragraph{Product Construction.}
Consider MDP $\mathcal{M}$, pRM $\mathcal{A}_R$, and let $\Delta$ be a set of proposition symbols for labeling regions in $X$ by function $L$. We define the synchronous product $\mathcal{M}\otimes\mathcal{A}_R$, which is expressed similarly to Definition \ref{mdp} as the following tuple $(X\times\Omega,U,r',\mathbb{P}',\lambda)$. For any state $(x_k,\tilde\varrho_k)\in X\times\Omega$, action $u_k\in U$ at time-step $k$ and observed proposition $\varphi_k=L(x_{k+1})\in2^\Delta$ from $\mathcal{M}$, the transition probability  $\mathbb{P}'\big[(x_{k+1},\tilde\varrho_{k+1})~|~(x_{k},\tilde\varrho_{k}),u_k\big]$ is as follows: 
\begin{equation}
    \label{pprime} 
    \begin{cases}
        \mathbb{P}[x_{k+1}|x_k,u_k]&\text{if }\tilde\varrho_{k}\notin \Omega_F\text{, }\tilde\varrho_{k+1}=\delta_\varrho(\tilde\varrho_{k},\varphi_k)\\
        \mathbb{P}[x_{k+1}|x_k,u_k]&\text{if }\tilde\varrho_{k}\in \Omega_F\text{, }\tilde\varrho_{k+1}=\tilde\varrho_{k}\\
        0&\text{ otherwise, }
    \end{cases}
\end{equation}
and the reward function $r':(X\times\Omega)\times U\times(X\times\Omega)\rightarrow \mathbb{R}$, which is an extension of $r$ over $X\times\Omega$, is defined as 
\begin{equation}
    \label{rprime}
    r'\big((x_{k},\tilde\varrho_{k}),u_k,(x_{k+1},\tilde\varrho_{k+1})\big)=\delta_r(\tilde\varrho_{k},\varphi).
    \end{equation}

Therefore, it is evident that $\mathcal{M}\otimes\mathcal{A}_R$ is an MDP. 
Hence, in the case where $\mathcal{M}\otimes\mathcal{A}_R$ is finite, the QL algorithm is applicable (cf. Alg. \ref{alg1} without lines $9$ to $11$). 
Otherwise, when computational resources are sufficient and the discretization of the infinite MDP is feasible, one can reduce the problem to a finite case where the QL algorithm can be applied. 
However, many real-world problems involve high-dimensional spaces where discretization is intractable. 
Moreover, practical control problems frequently require operations in continuous state spaces with continuous actions to achieve complex tasks specified by infinite pRMs. These scenarios are beyond the scope of the QL algorithm. To address these challenges, we employ the DDPG algorithm (cf. Alg. \ref{alg2}, excluding lines $11$ to $13$ in the Appendix), an off-the-shelf RL approach, enabling an RL agent to learn policies over  infinite MDPs and pRMs.

\section{Reinforcement Learning with pRMs}
\label{exploiting_rm}
Icarte et al. demonstrated how the notion of RM can be exploited through the generation of synthetic experiences and the use of potential-based reward shaping \cite{ng1999policy} over RMs to achieve efficient learning. We extend these ideas in the context of pRMs.

\subsection{Counterfactual Experiences}
Consider a pRM $\mathcal{A}_R$ and an \emph{actual} experience $(x,\tilde\varrho,u,r,x',\tilde\varrho')\in X\times\Omega\times U\times\mathbb{R}\times X\times\Omega$ based on the observation of the environment $(x,u,x')\in X\times U\times X$, where the agent evolves from state $(x,\tilde\varrho)$ to $(x',\tilde\varrho')$ on $\mathcal{M}\otimes\mathcal{A}_R$, and obtains reward $r$ by taking action $u$. So, in each training step, we enable the agent to access some pRM experiences (pRME) generated at each non-terminal state of $\mathcal{A}_R$ in $\Omega\setminus\Omega_F$, which are gathered in the following set
\begin{equation}
\big\{\big( x, \tilde\varrho, u, \delta_r(\tilde\varrho, L(x')), x', \delta_\varrho(\tilde\varrho, L(x'))\big)~\big|
        ~\forall\tilde\varrho \in \Omega \setminus \Omega_F\big\}\subseteq X {\times} \Omega {\times} U {\times} \mathbb{R} {\times} X {\times} \Omega.
        \label{pgrm}
\end{equation}

Note that due to the continuous state nature of the pRM, an infinite number of counterfactual experiences can, in principle, be generated according to \eqref{pgrm}. To make this computationally tractable, a finite subset of these experiences may be selected using random sampling or prioritized sampling with a weighted scheme. In this work, we adopt uniform random sampling to extract a finite number of experiences from those generated in \eqref{pgrm}, and make these available to the learning agent during training. Consequently, the agent can learn the correct behaviour at each pRM state more rapidly through the reuse of the counterfactual experiences in \eqref{pgrm}. This mechanism, enables the agent to receive different rewards based on the state of $\mathcal{A}_R$ at any given time. 

We note that, given an actual experience, the counterfactual experiences generated according to \eqref{pgrm} not only subsume those generated for RMs in \cite{icarte2022reward}, which are uniquely and deterministically determined, but also enhance the set of pRMEs. By incorporating \eqref{prm_ode} into pRMs, the pRMEs are improved both in cardinality and quality. Specifically, unlike the completely counterfactual experiences in \cite{icarte2022reward}, the physical information encoded in \eqref{pgrm} provides meaningful guidance based on the dynamics, ensuring that pRMEs are not entirely fake experiences but rather incorporate realistic intelligence to facilitate learning. This highlights the crucial need to exploit pRMEs (\emph{e.g.,} cf. line $9$ to $11$ of Alg. \ref{alg1}). 

Next, we present algorithms for using pRME in both finite and continuous settings using QL and DDPG.

\paragraph{QL on finite MDPs and pRMs with pRMEs.} We adapt pRME to QL by incorporating the set of experiences from \eqref{pgrm} into the actual experiences, creating a pool of experiences. The agent then uses this pool during each training iteration to update the state-action values function Q. The pseudocode is provided in Alg. \ref{alg1}.
\begin{algorithm}[t]
	\caption{QL on finite MDPs exploiting pRMs 
    }
 \label{alg1}
 \begin{algorithmic}[1]
\REQUIRE $X$, $U$, $\Delta$, $L$, $\Omega$, $\Omega_F$, $\Omega_0$, $\delta_\varrho$, $\delta_r$ and $\lambda,\varepsilon,\kappa\in(0,1)$
\STATE Initialize values for $\tilde q(x,\tilde\varrho,u)$, $\forall(x,\tilde\varrho,u)\in X\times \Omega\times U$
\WHILE{the training steps is less than some threshold}
\STATE Initialize $(x,\tilde\varrho)\leftarrow (x_0,\tilde\varrho_0)$
\WHILE{$\tilde\varrho\notin\Omega_F$}
\STATE Select action $u\in U$ on $(x,\tilde\varrho)$ based on policy (\emph{e.g.,} $\varepsilon$-greedy) obtained from $\tilde q$ 
\STATE Observe the succeeding state $x'$ on taking action $u\in U$ and initialize set of experiences $\mathcal{E}\leftarrow\emptyset$
\STATE Update pRM next state $\tilde\varrho'\leftarrow\delta_\varrho(\tilde\varrho,L(x'))$ using \eqref{prm_dyn} with reward $r\leftarrow \delta_r(\tilde\varrho,L(x'))$ using \eqref{prm_rew}
\STATE Update $\mathcal{E}\leftarrow\{(x,\tilde\varrho,u,r,x',\tilde\varrho')\}$
\IF{pRME is involved}
\STATE Update $\mathcal{E}$ according to \eqref{pgrm}
\ENDIF
\FOR{$(x,\hat\varrho,u,\hat r,x',\hat\varrho')\in\mathcal{E}$}
\IF{$\hat\varrho'\notin\Omega_F$}
\STATE $\tilde q(x,\hat\varrho,u)\stackrel{\kappa}{\leftarrow}\hat r+\lambda\max_{u'\in U}\{\tilde q(x',\hat\varrho',u')\}$
\ELSE
\STATE $q(x,\hat\varrho,u)\stackrel{\kappa}{\leftarrow}\hat r$
\ENDIF
\ENDFOR
\STATE Update state $(x,\tilde\varrho)\leftarrow(x',\tilde\varrho')$
    \ENDWHILE
    \ENDWHILE
\end{algorithmic}
\end{algorithm}
The next result, proved in the Appendix, 
establishes the convergence of Alg. \ref{alg1}.
\begin{theorem}
    \label{ql_convg_thrm}
    Consider an unknown finite MDP $\mathcal{M}$. Let $\mathcal{A}_R$ be a finite pRM as in Definition \ref{rm}. If every state-action pair $\big(x,\tilde\varrho,u\big)\in X\times\Omega\times U$ is visited infinitely often, then Alg. \ref{alg1} converges to the optimal policy asymptotically.
\end{theorem}
\paragraph{DDPG on infinite MDPs and pRMs with pRMEs.}
We leverage pRMEs in DDPG by incorporating experiences in \eqref{pgrm} into the replay buffer $\mathcal{B}$, from which samples are drawn for learning. The complete procedure is systematically detailed in Algorithm~\ref{alg2} in the Appendix.

\begin{remark}
\label{comb_prm_rmk}
Note that complex tasks may require multiple pRMs. Therefore, we can specify $T\in\mathbb{N}_{\ge1}$ distinct tasks, each associated with a different pRM $\mathcal{A}_{R_i}$, where $i\in[1;T]$. At time-step $k$, the assigned reward for task $i$ is represented as $r_{i_k}$, determined by the reward structure embedded in $\mathcal{A}_{R_i}$, according to Definition \ref{rm}. Thus, in a training episode, the total reward at time-step $k$ across all $T$ pRMs is given by $\frac{1}{T}\sum_{i=1}^T r_{i_k}$, which provides more interpretable results over a discounted reward approach. The average reward method avoids the issue of sub-optimal steps, which can be disproportionately penalized in reward structure exploitation schemes like pRME, due to their dependence on the discount factor.
\end{remark}

\subsection{Reward Shaping}

Consider the pRM $\mathcal{A}_{R2}$ in Figure~\ref{fig_rm2}, which encodes the task structure for the environment in the motivating example~\ref{mot_ex}. If only $\zeta_1$ yields a non-zero reward, the agent is rewarded only after completing all tasks, resulting in a highly sparse reward structure that makes learning difficult. Introducing intermediate rewards can mitigate this challenge by guiding the agent through sub-goals. However, such shaping must preserve the set of optimal policies under the original reward structure. Given a pRM $\mathcal{A}_{R}=(\Omega,\Omega_F,\Omega_0,\delta_\varrho, \delta_r)$, we can construct an MDP $\mathcal{M}_{\mathcal{A}_{R}}=(X_{\mathcal{A}_{R}},U_{\mathcal{A}_{R}},r_{\mathcal{A}_{R}},\mathbb{P}_{\mathcal{A}_{R}},\lambda)$, where $X_{\mathcal{A}_{R}}=\Omega$, $U_{\mathcal{A}_{R}}=2^\Delta$, and $\lambda\in(0,1)$. For any $\varphi\in2^\Delta$ and $\tilde\varrho,\tilde\varrho'\in\Omega$, the reward function $r_{\mathcal{A}_{R}}$ and transition probability $\mathbb{P}_{\mathcal{A}_{R}}$ are defined as follows:
\begin{align}
r_{\mathcal{A}_{R}}(\tilde\varrho,\varphi,\tilde\varrho')&=\delta_r(\tilde\varrho,\varphi)\text{ if }\tilde\varrho\in\Omega\setminus\Omega_F\text{ else }0,\label{rs_eq1}\text{ and}\\
\mathbb{P}_{\mathcal{A}_{R}}[\tilde\varrho'~|~\tilde\varrho,\varphi]&=\begin{cases}
    1&\text{if }\tilde\varrho\in\Omega_F\text{, }\tilde\varrho=\tilde\varrho',\\
    1&\text{if }\tilde\varrho\in\Omega\setminus\Omega_F\text{, }\tilde\varrho'=\delta_\varrho(\tilde\varrho,\varphi),\\
    0&\text{otherwise.}
\end{cases}\label{rs_eq2}
\end{align}
Thus, we establish the following result, based on \cite[Theorem 1]{ng1999policy}, which formally guarantees the preservation of optimal policies in MDPs with shaped rewards. 
The proof follows a similar approach as in the original version.
\begin{theorem}[Reward Shaping (RS)]
    \label{rs_correct}
    Suppose $\mathcal{A}_{R}=(\Omega,\Omega_F,\Omega_0,\delta_\varrho, \delta_r)$ is a pRM as in Definition \ref{rm}. Let the MDP $\mathcal{M}_{\mathcal{A}_{R}}=(X_{\mathcal{A}_{R}},U_{\mathcal{A}_{R}},r_{\mathcal{A}_{R}},\mathbb{P}_{\mathcal{A}_{R}},\lambda)$ be constructed according to \eqref{rs_eq1} and \eqref{rs_eq2}, with $\lambda\in(0,1)$. For any $\varphi\in2^\Delta$, $\tilde\varrho,\tilde\varrho'\in\Omega$, and a function $\Phi:X_{\mathcal{A}_{R}}\rightarrow\mathbb{R}$, let 
    \begin{equation}
        \label{rs_function}
r_{\mathcal{A}_{R}}'(\tilde\varrho,\varphi,\tilde\varrho')=r_{\mathcal{A}_{R}}(\tilde\varrho,\varphi,\tilde\varrho')-\lambda\Phi(\tilde\varrho')+\Phi(\tilde\varrho)
    \end{equation}
    be a potential-based reward shaping function. Then the set of optimal policies for MDP $\mathcal{M}_{\mathcal{A}_{R}}'=(X_{\mathcal{A}_{R}},U_{\mathcal{A}_{R}},r_{\mathcal{A}_{R}}',\mathbb{P}_{\mathcal{A}_{R}},\lambda)$ coincides with that of $\mathcal{M}_{\mathcal{A}_{R}}$.
\end{theorem}

Note that, in \eqref{rs_function}, the potential function takes a negative value to encourage the agent to move toward pRM states associated with task completion. We derive the potential function by solving the deterministic MDP $\mathcal{A}_{R}$ using value iteration (cf. Appendix). Alternative approaches can be found in \cite{post2015simplex,bertram2018fast}.  

\begin{figure*}[t]
    \centering
    \begin{subfigure}{0.32\textwidth}
        \centering
        \includegraphics[width=\textwidth]{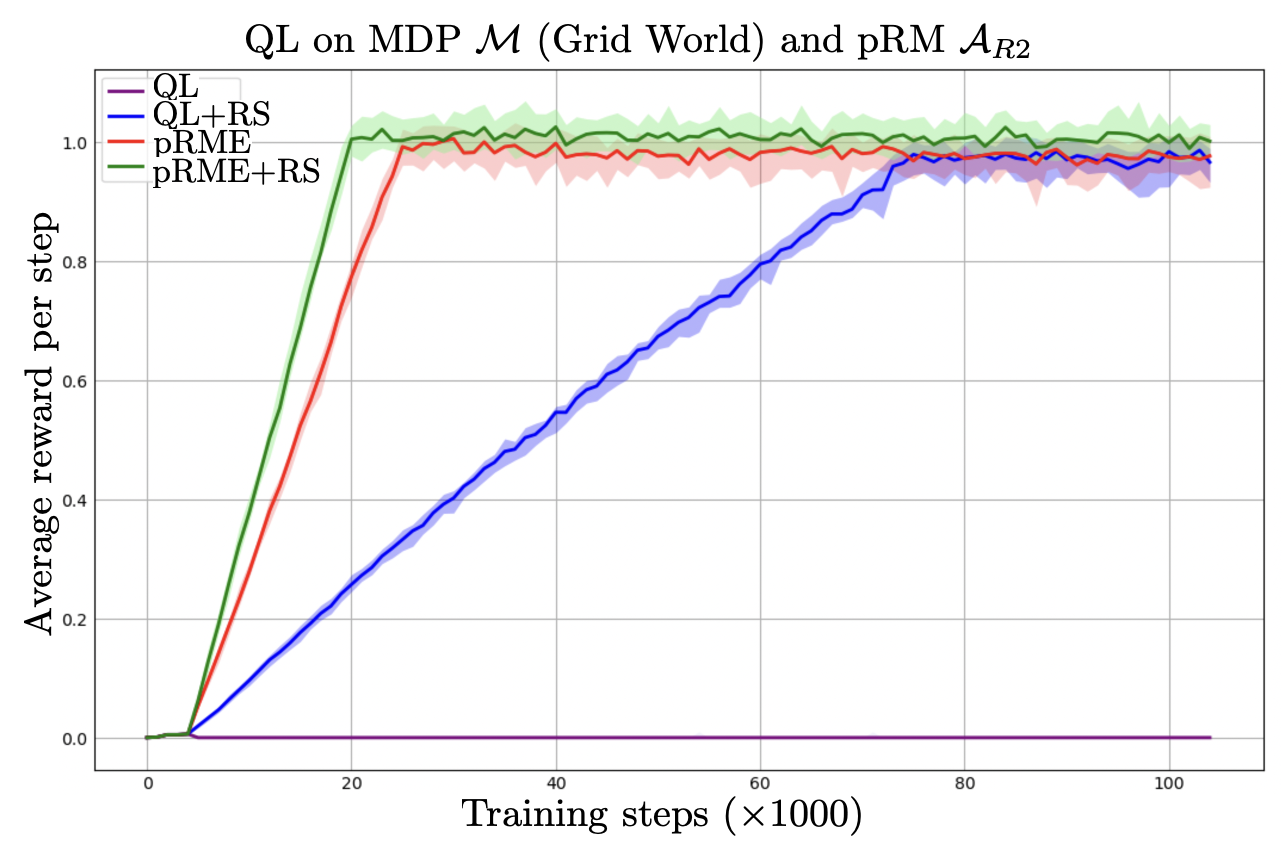}
        \caption{}
        \label{fig_gridworld_rew}
    \end{subfigure}
    \begin{subfigure}{0.32\textwidth}
        \raggedleft
        \includegraphics[width=\textwidth]{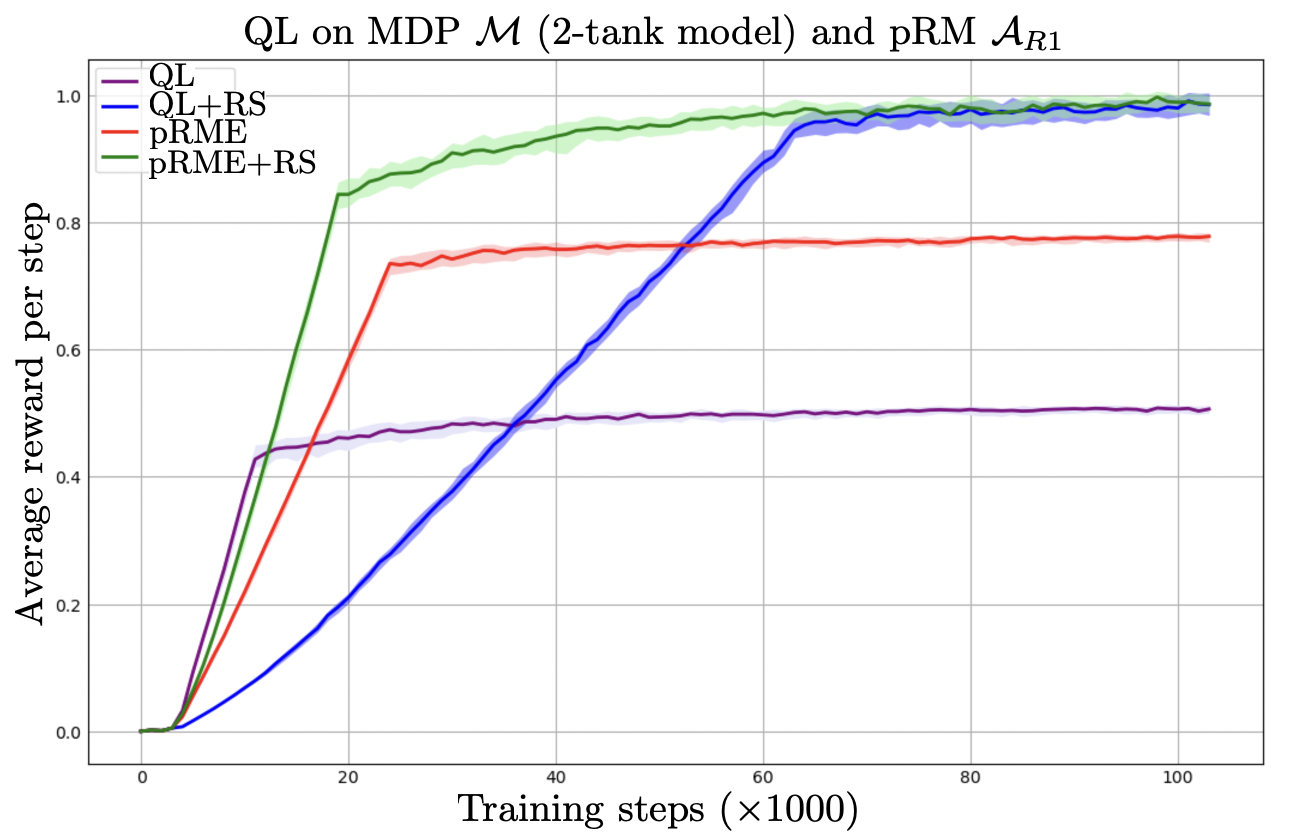}
        \caption{}
        \label{fig_2tank_rew}
    \end{subfigure}
    \begin{subfigure}{0.32\textwidth}
        \raggedleft
        \includegraphics[width=\textwidth]{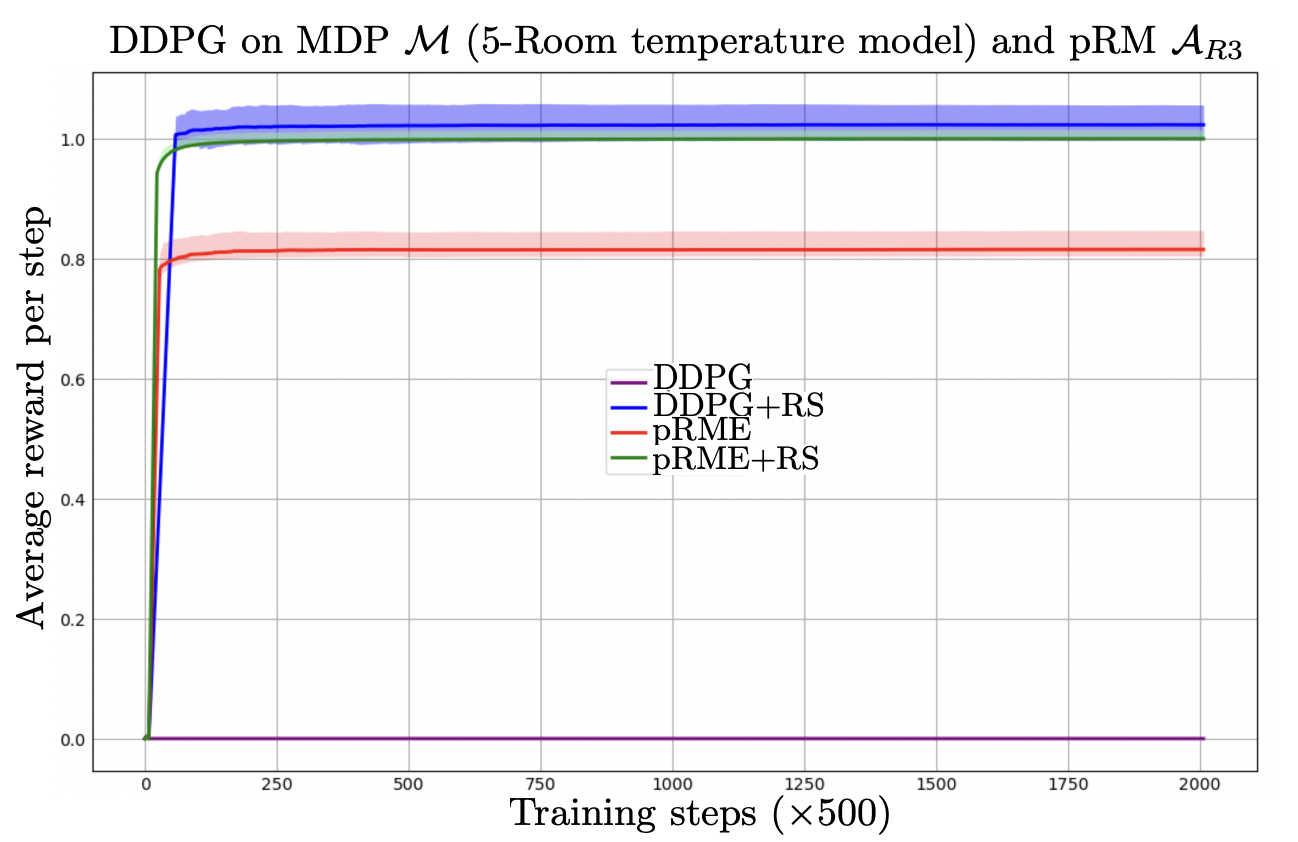}
        \caption{}
        \label{fig_5rm_rew}
    \end{subfigure}
    \\[1.5ex]
    \begin{subfigure}{0.32\textwidth}
        \raggedleft
        \includegraphics[width=\textwidth]{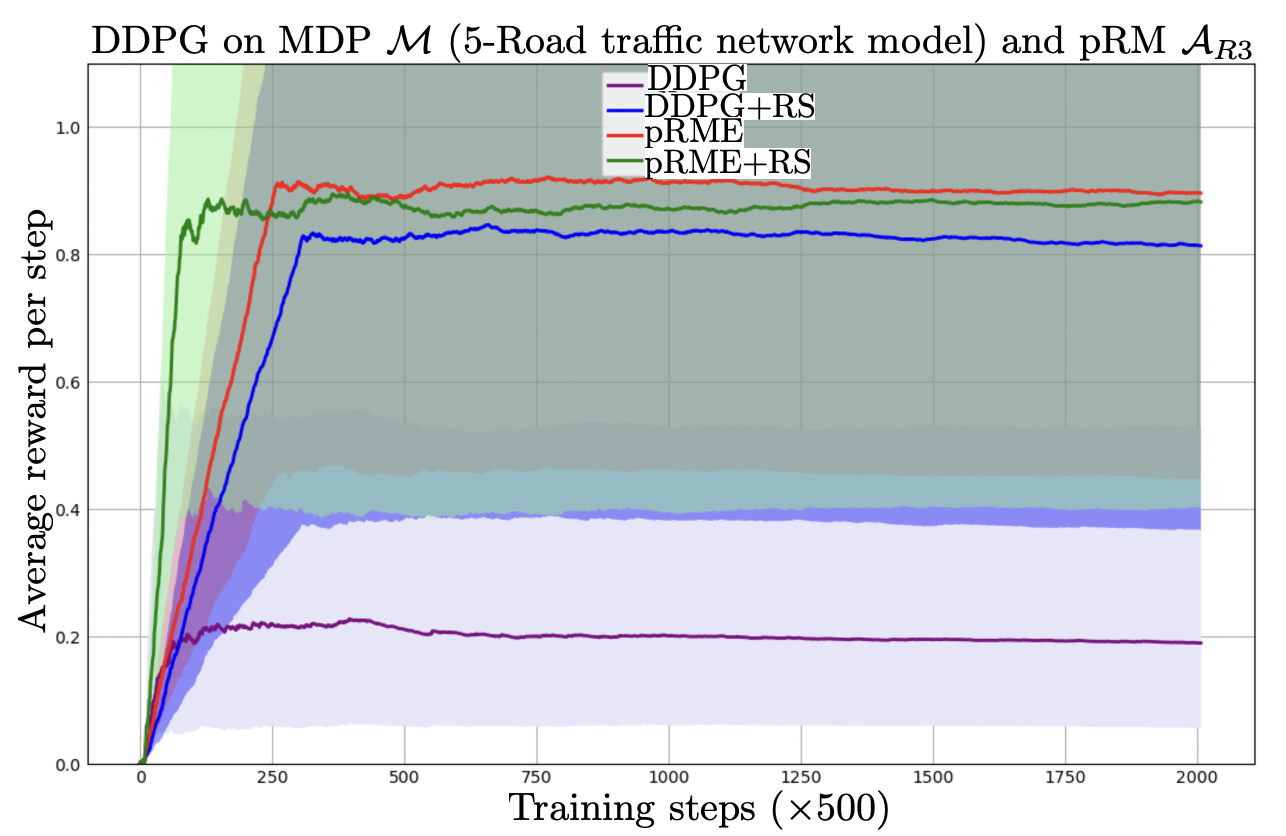}
        \caption{}
        \label{fig_5rd_net_rew}
    \end{subfigure}
    \begin{subfigure}{0.32\textwidth}
        \raggedleft
        \includegraphics[width=\textwidth]{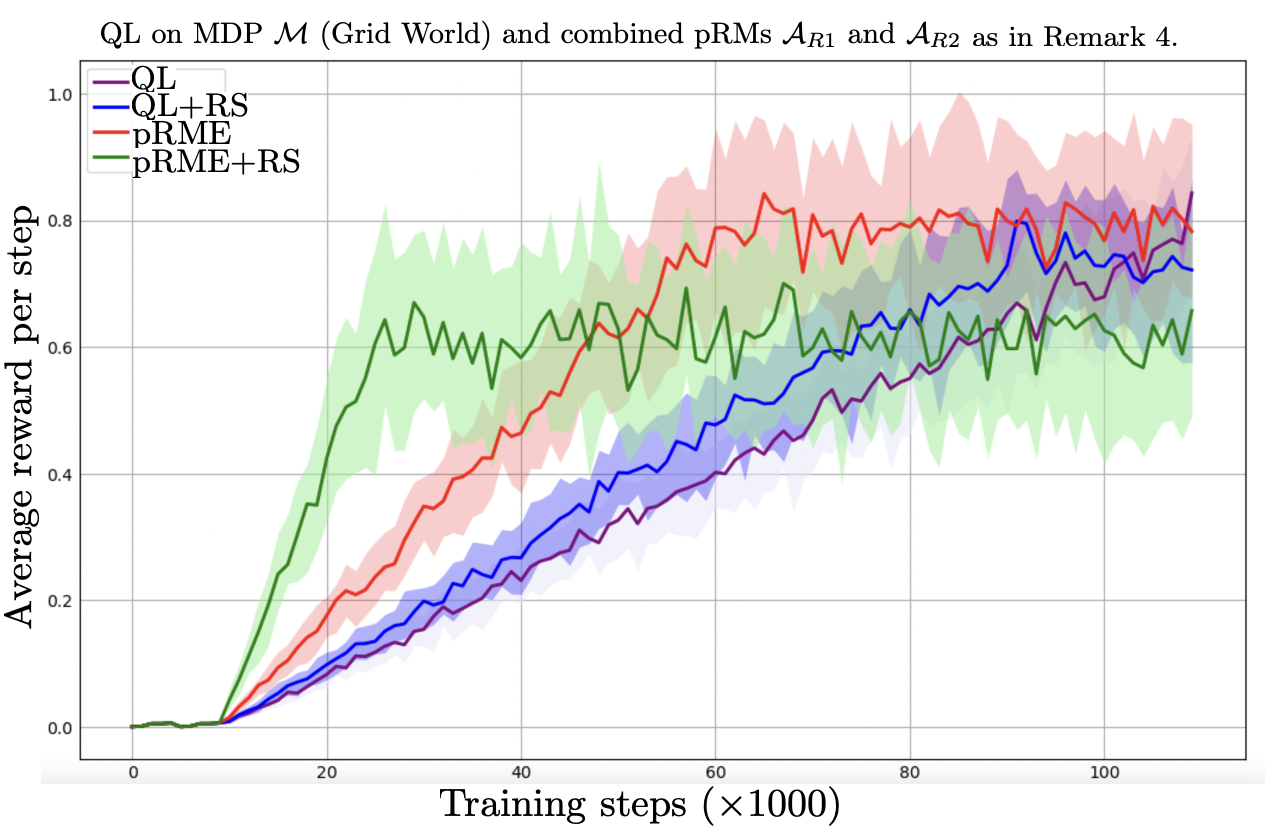}
        \caption{}
        \label{fig_multi}
    \end{subfigure}
    \begin{subfigure}{0.32\textwidth}
        \raggedleft
        \includegraphics[width=\textwidth]{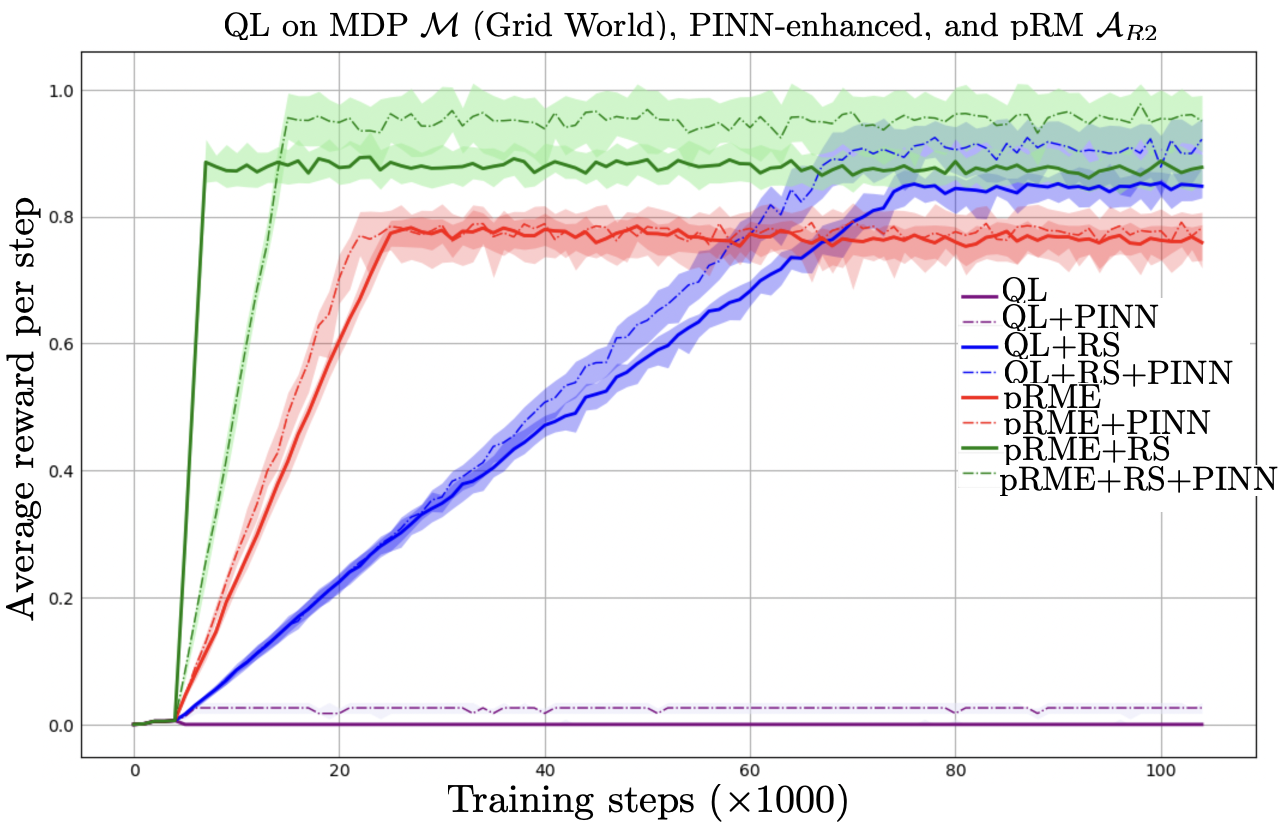}
        \caption{}
        \label{fig_abla}
    \end{subfigure}
    \caption{ The plots show the average reward per training step for the RL agent learning tasks specified by $\mathcal{A}_{R2}$ in (a), $\mathcal{A}_{R1}$ in (b), both pRMs for multi-tasks in (e), and $\mathcal{A}_{R3}$ in (c) and (d). The plot (f) is for tasks specified by $\mathcal{A}_{R2}$, with the results obtained from Alg. \ref{alg1} applied to the Office World environment enhanced with PINN. These experiments utilize Alg. \ref{alg1} and Alg. \ref{alg2}, applied to their respective environments.}
    \label{fig_exp_result}
\end{figure*}

\section{Experiments}
\label{case_study}
We experimentally demonstrate the expressiveness of pRMs in both finite and continuous stochastic physical environments. Using the off-policy RL algorithms in Algorithms~\ref{alg1} and~\ref{alg2}, we show that incorporating pRM reward structures accelerates learning across diverse control tasks. We do not adopt hierarchical RL for RMs, as it often yields suboptimal policies in the limit~\cite{icarte2022reward}.

The environments include an office gridworld, a 2-tank system, a 5-room temperature model, and a 5-road traffic network, along with PINN-enhanced variants. Full experimental details, including hyperparameters and implementation specifics, are in the Appendix.

\paragraph{Qualitative Results.}
Figure \ref{fig_exp_result} illustrates faster reward maximization with both Alg. \ref{alg1} and Alg. \ref{alg2}, even in multi-pRM settings, as shown in Figure \ref{fig_multi}. This demonstrates quicker policy stabilization, yielding speedy convergence in pRM-enhanced approaches. The performance is generally improved on combining pRMEs with reward shaping (RS), yielding results that surpass those achieved using pRMEs alone. Moreover, policy interpretability is significantly enhanced with pRMs. For instance, in the Office World experiment, the coffee temperature dynamics in pRM $\mathcal{A}_{R2}$ provide the agent with prior physical knowledge, systematically integrated into both the task and reward function, making the agent aware of physical constraints and objectives.

\begin{figure}[t!]
        \centering
        \includegraphics[width=1.0\linewidth]{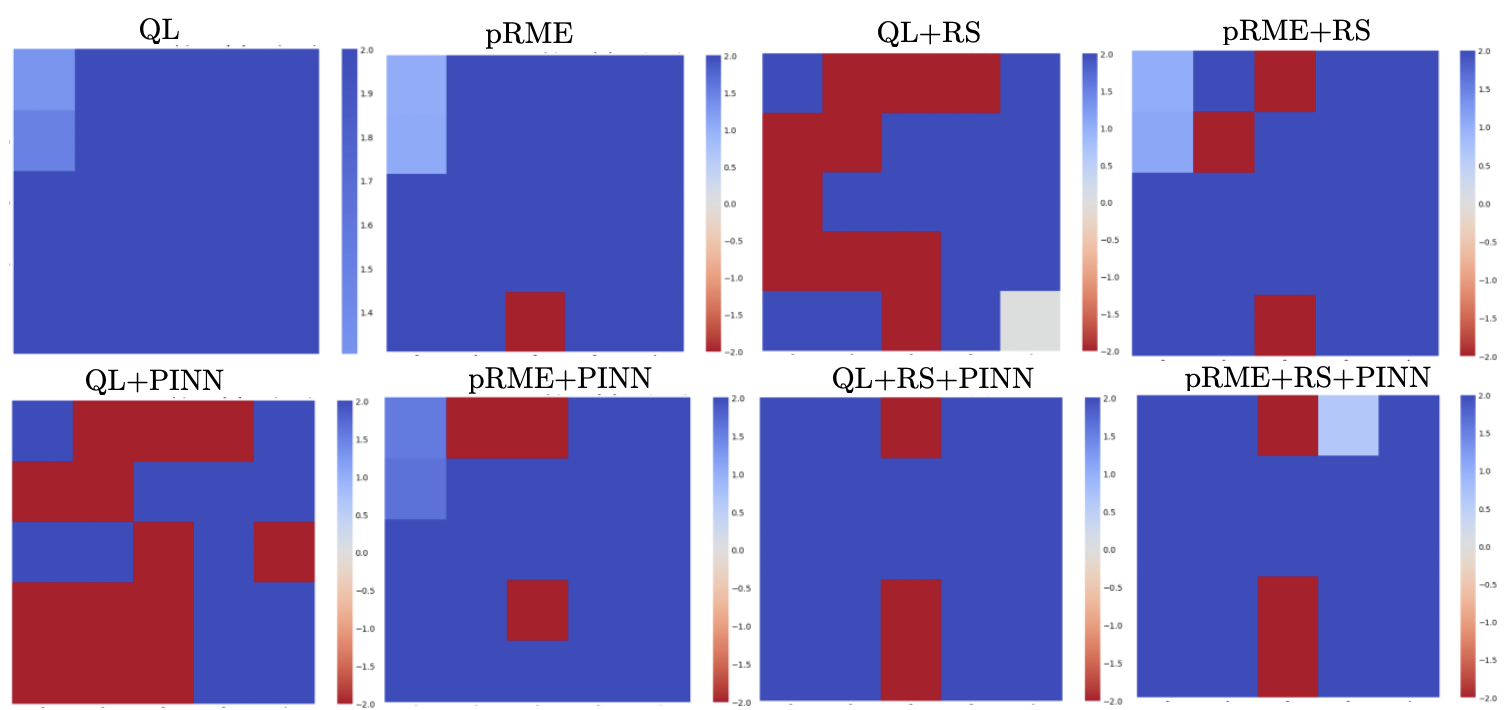}
    \caption{The heatmaps show safe (blue) and unsafe (red) states, with intensity corresponding to the maximum Q-values based on the  objectives defined by pRM $\mathcal{A}_{R2}$.}
    \label{ab}
\end{figure}

\paragraph{Quantitative Results.}
Figures \ref{fig_exp_result} demonstrate a significant improvement in cumulative rewards per training step with pRM-enhanced RL. The pRME and pRME+RS approaches outperform QL+PINN in terms of rewards. However, both PINN and pRM-enhanced methods show notably better results (cf. Figure \ref{fig_abla} and \ref{ab}), suggesting that further exploration of PINN could bolster the pRM approach in RL. Figure \ref{fig_multi} shows that averaging rewards from two pRMs (see Remark \ref{comb_prm_rmk}) can dilute task-specific signals—especially when reward shaping is applied independently—leading to conflicting feedback and suboptimal performance. In contrast, approaches like QL with tailored rewards or pRME provide more focused guidance. Notably, combining pRMs ensures the learned policy is, at worst, suboptimal, unlike single-pRM cases where methods that do not exploit the pRM perform poorly. Additionally, the number of episodes required for convergence is reduced in the pRM-enhanced methods compared to the baseline approaches, promoting sample efficiency during learning. Consequently, these baselines exhibit a higher number of violations (e.g., collisions with flowers or delivering cold coffee in Example \ref{mot_ex}), highlighting unsafe behaviors compared to pRM-guided learning, as also reflected in the classification in Figure \ref{ab}. Furthermore, incorporating pRME increases the complexity of both algorithms, as shown by the runtimes in Table \ref{tab1}. This can be mitigated through parallelization and deployment on more powerful GPU hardware. Despite this trade-off, pRME proves valuable, as it enables more efficient sampling and reduces the inherent reward sparsity in pRM through RS, allowing the RL agent to find policies and maximize rewards more quickly than without pRME and reward shaping.

\begin{table}[t!]
\centering
\resizebox{0.6\textwidth}{!}{
\begin{tabular}{c|c|c||c|c|c} 
\hline
\shortstack{Alg. $1$\\(in seconds)}& Fig.~\ref{fig_gridworld_rew} & Fig.~\ref{fig_2tank_rew} &\shortstack{Alg. $2$\\(in minutes)}& Fig.~\ref{fig_5rm_rew}  &Fig.~\ref{fig_5rd_net_rew} \\ 
\hline
QL&$2.6\pm0.1$ & $4.2\pm0.7$ &DDPG & $2.2\pm0.3$ &$0.3\pm0.1$\\
\hline
{\small QL+RS}&$3.8\pm0.3$ & $6\pm1.5$ &{\small DDPG+RS} & $43\pm0.7$ &$50.4\pm0.1$ \\
\hline
pRME&$14\pm3.9$ &$20\pm2.1$ &pRME &$53\pm0.6$ &$69\pm5.5$\\
\hline{\small pRME+RS}&$26\pm1.8$&$24\pm1.9$&{\small pRME+RS} &$123\pm1.3$ &$131\pm3.3$\\
\hline
\end{tabular}
}
\caption{Average runtime per trial for the experimental results shown in Figure~\ref{fig_exp_result}.}
\label{tab1}
\end{table}

\paragraph{Ablation Study.}
Without the pRM physical dynamics, considering Example \ref{mot_ex}, the learned policy lacks the coffee temperature dynamics, making the agent less task-guided and absent of physical constraints. This results in a less robust policy compared to those learned with pRM-enhanced methods, as shown in the safe behavior heatmap in Figure \ref{ab}. Figure \ref{fig_abla} illustrates the superior performance of pRM-enhanced approaches in the multi-task setting, where we combined tasks from pRMs $\mathcal{A}_{R1}$ and $\mathcal{A}_{R2}$ over Example \ref{mot_ex} following Remark \ref{comb_prm_rmk}.

Furthermore, the experimental results from \cite{icarte2022reward} on the Office World example, has its fastest RL method reaches maximum rewards after approximately $20000$ training steps without the physics component in the pRM, as outlined in \eqref{prm_ode}. In contrast, the pRM-enhanced RL scheme attains maximum rewards in under $5000$ training steps. This demonstrates that incorporating physics-informed guidance into both the task and reward structure not only accelerates learning but also promotes safer behavior as evident in Figure \ref{ab}. 

Additionally, Figure \ref{fig_abla} highlights the impact of PINNs, which contribute to better reward maximization and faster convergence when integrated into the pRM-enhanced RL scheme. This underscores the advantages of fully integrating PINNs with pRMs, as evidenced by the safe behavior shown in Figure \ref{ab}. However, it is worth noting that while pRM-explored approaches perform well on their own, using only PINN without the task-driven physics constraint results in lower reward acquisition, although it still outperforms the baseline approaches.

\section{Conclusion}
\label{conclusion}
We introduced pRMs as a symbolic, expressive framework for specifying structured reward functions and control tasks in RL. pRMs support infinite regular languages and temporally extended behaviors with arbitrarily sparse rewards, allowing compact representation of complex task specifications. By leveraging pRM structure, RL agents achieve improved learning efficiency, reduced sample complexity, and higher reward performance---especially in settings with limited agent-environment interaction, as demonstrated in our case studies. 
Inspired by hybrid automata, pRMs integrate formal task specifications with physical structure to enable sample-efficient and explainable learning. They abstract high-level observations into structured rewards, guiding exploration and policy learning. Unlike standard hybrid automata, which focus on binary satisfaction, pRMs shape rewards from continuous dynamics, even when the agent remains in the same state---a capability absent in traditional reward machines. While PINNs ensure that environment observations conform to physical laws, pRMs use those observations to shape rewards and drive learning. Together, PINNs enhance the physical fidelity of observations, while the hybrid structure of pRMs increases expressiveness beyond standard RMs. This work assumes access to a perfect labeling function that maps transitions to pRM events---standard in RM literature. Future directions include handling noisy labels, learning pRMs from demonstrations, and generalizing the dynamics in \eqref{prm_ode} to (stochastic) PDEs~\cite{da2014stochastic}, potentially using PINN- or PIRL-based methods. These extensions could further enable physics-aware, high-level task learning within a unified system.

\bibliographystyle{alpha}
\bibliography{biblio.bib}

\appendix

\section{Proof of Theorem \ref{ql_convg_thrm}}
\begin{proof}
    The proof is established as follows. Suppose that the pRMEs in \eqref{pgrm} are contained in a set $\hat{\mathcal{E}}$, and are sampled according to some transition probability distribution $\hat{\mathbb{P}}$. Let $\mathbb{P}_{\mathcal{A}}$ be the coupling probability distribution of $\hat{\mathbb{P}}$ and $\mathbb{P}$ over the set of experiences $\mathcal{E}$ in Algorithm \ref{alg1}. Then by the coupling inequality result in \cite{den2012probability}, it holds that for any $(\hat x,\tilde\varrho,u,\hat r,\hat x',\tilde\varrho')\in\hat{\mathcal{E}}$ and $(x,\tilde\varrho,u,r,x',\tilde\varrho')\in\mathcal{E}\setminus\hat{\mathcal{E}}$, such that $\hat{\mathbb{P}}[\hat x'~|~\hat x,u]$, $\mathbb{P}[x'~|~x,u]$ are valid probabilities, then
\begin{equation}
        2\mathbb{P}_{\mathcal{A}}\big[(x',\hat x',\tilde\varrho)~|~(x,\hat x,\tilde\varrho),u\big]\ge\big\lvert\mathbb{P}\big[x'~|~x,u\big]-\hat{\mathbb{P}}\big[\hat x'~|~\hat x,u\big]\big\rvert.
    \end{equation}
    Therefore, we define the probability $\mathbb{P}_{\mathcal{A}}[(x',\hat x',\tilde\varrho)~|~(x,\hat x,\tilde\varrho),u]$ over $\mathcal{E}$ as follows:
    \begin{equation}
    \label{p_couple} 
    \begin{cases}
        \mathbb{P}[(x',\tilde\varrho')~|~(x,\tilde\varrho),u]&\text{ if }x=\hat x\text{ and }x'=\hat x'\\
        \frac{1}{2}\big\lvert\mathbb{P}[x'~|~x,u]-\hat{\mathbb{P}}[\hat x'~|~\hat x,u]\big\rvert&\text{ if }x\neq\hat x\text{ and }x'\neq\hat x'\\
        0&\text{ otherwise, }
    \end{cases}
\end{equation}
where $\mathbb{P}[(x',\tilde\varrho')~|~(x,\tilde\varrho),u]$ is defined according to \eqref{pprime}. With the above transformation of the transition probability distribution for finite MDP $\mathcal{M}\otimes\mathcal{A}_R$, the remainder of the proof follows the same reasoning as the convergence of ordinary QL outlined in \cite{watkins1992q}.
\end{proof}

\section{DDPG on infinite MDPs and pRMs with pRMEs}
\begin{algorithm}[H]
	\caption{DDPG on infinite MDPs exploiting pRMs}
 \label{alg2}
 \begin{algorithmic}[1]
\REQUIRE $X$, $U$, $\Delta$, $L$, $\Omega$, $\Omega_F$, $\tilde\varrho_0$, $\delta_\varrho$, $\delta_r$ and $\lambda,\iota\in(0,1)$
\STATE Initialize actor network $\rho_\mu(x,\tilde\varrho)$ and critic network 
$\tilde q_\vartheta(x,\tilde\varrho,u)$ parameterized with weights $\mu$ and $\vartheta$, respectively, where $(x,\tilde\varrho,u)\in X\times \Omega\times U$
\STATE Initialize a replay buffer $\mathcal{B}\leftarrow\emptyset$ and target network parameters as 
$\mu'\leftarrow\mu$ and $\vartheta'\leftarrow\vartheta$ 
    \WHILE{the training steps is less than some threshold}
    \STATE Initialize a random process $\mathcal{R}$ to explore $U$
\STATE Observe an initial state $(x_0,\tilde\varrho_0)$ 
\FOR{time-step $k\in[0;\texttt{number\_of\_episode}]$ }
\STATE Select action $u_k=\rho_\mu(x_k,\tilde\varrho_k)+\varsigma_k$ where $\varsigma_k\sim\mathcal{R}$ and initialize set of experiences $\mathcal{E}\leftarrow\emptyset$
\STATE Observe the next state $x_{k+1}$ and reward \\$r_k\leftarrow \delta_r(\tilde\varrho_k,L(x_{k+1}))$ on taking action $u_k$
\STATE Using \eqref{prm_dyn}, update $\tilde\varrho_{k+1}\leftarrow\delta_\varrho(\tilde\varrho_k,L(x_{k+1}))$ 
\STATE Update $\mathcal{B}\leftarrow\{(x_k,\tilde\varrho_k,u_k,r_k,x_{k+1},\tilde\varrho_{k+1})\}$
\IF{pRME is involved}
\STATE Update $\mathcal{E}$ according to \eqref{pgrm}, and $\mathcal{B}\leftarrow\mathcal{B}\cup\mathcal{E}$
\ENDIF
\STATE Sample a mini-batch of $N$ experiences $\tilde{\mathcal{E}}$ from $\mathcal{B}$
\FOR{$(x,\hat\varrho,u,\hat r,x',\hat\varrho')\in\tilde{\mathcal{E}}$}
\IF{$\hat\varrho'\notin\Omega_F$}
\STATE $y(x')\leftarrow\lambda\tilde q_{\vartheta'}(x',\hat\varrho',\rho_\mu(x_{k+1},\tilde\varrho_{k+1}))+$ $\hat r$
\ELSE
\STATE $y(x')\leftarrow\hat r$
\ENDIF
\ENDFOR
\STATE Minimize 
$\frac{1}{N}\sum\limits_{\tilde{\mathcal{E}}}\big(y(x'){-}\tilde q_\vartheta(x,\hat\varrho,u)\big)^2$ to update $\tilde q_\vartheta$
\STATE Compute one-step of gradient ascent utilizing $\frac{1}{N}\nabla_\mu\Sigma_{\tilde{\mathcal{E}}}q_\vartheta(x,\hat\varrho,\rho_\mu(x,\hat\varrho))$ to update the actor $\rho_\mu$
\STATE Update $\vartheta'\stackrel{\iota}{\leftarrow}\vartheta$ and $\mu'\stackrel{\iota}{\leftarrow}\mu$
    \ENDFOR
    \ENDWHILE
\end{algorithmic}
\end{algorithm}

\section{Algorithm for value iteration}
The value iteration procedure for determining the potential function $\Phi$ in Theorem \ref{rs_correct} is outlined below:

\begin{algorithm}[H]
	\caption{Value iteration for potential-based RS}
 \label{alg3}
 \begin{algorithmic}[1]
\REQUIRE $\Omega$, $\Omega_F$, $\Delta$, $\delta_\varrho$, $\delta_r$ and $\lambda,\texttt{tol}\in(0,1)$
\FOR{$\tilde\varrho\in\Omega$}
\STATE Initialize state potential value $\Phi(\tilde\varrho)\leftarrow0$
\ENDFOR
\STATE $\texttt{err}\leftarrow\texttt{tol}$
\WHILE{$\texttt{err}\ge\frac{1-\lambda}{\lambda}\texttt{tol}$}
\STATE $\texttt{err}\leftarrow0$
\FOR{$\tilde\varrho\in\Omega\setminus\Omega_F$}
\STATE $\Phi'\leftarrow\max_{\varphi\in2^\Delta}\{\delta_r(\tilde\varrho,\varphi)+\lambda\Phi(\delta_\varrho(\tilde\varrho,\varphi))\}$
\STATE $\texttt{err}=\max\{\texttt{err},|\Phi'-\Phi(\tilde\varrho)|\}$
\STATE $\Phi(\tilde\varrho)\leftarrow\Phi'$
    \ENDFOR
    \ENDWHILE
    \ENSURE $\Phi$
\end{algorithmic}
\end{algorithm}

\section{Related Work} 
\label{related_wk}
The formal specification of tasks for learning agents and the use of structured reward systems have gained increasing attention. In particular, the use of formal languages to define reward structures has seen a notable rise. (\emph{e.g.,} \cite{icarte2022reward} and the references therein). In recent years, researchers have begun developing data-driven policy synthesis techniques aimed at satisfying temporal properties \cite{baier2008principles}. A substantial body of literature exists on safe RL (\emph{e.g.,} \cite{garcia2015comprehensive,recht2019tour,efroni2020exploration}). In particular, the results in \cite{efroni2020exploration} guarantee minimized violations of certain unknown constraints in MDPs, while the agent maximizes cumulative rewards. Recent studies have focused on learning policies that maximize the satisfaction probability of temporal properties using discounted RL \cite{sadigh2014learning,hasanbeig2023certified,bozkurt2020control,brazdil2014verification,fu2014probably,hasanbeig2019reinforcement,oura2020reinforcement}. The results in \cite{lavaei2020formal,lavaei2023compositional} focus on transferring the probability of satisfying a finite-horizon property, learned through RL techniques on a control system's abstraction, to the original system.

The concept of a reward machine (RM) was first introduced in \cite{icarte2018using} as a type of finite-state machine designed to specify reward structures for RL agents. In addition, there have been several advances in learning such finite-state machines, particularly when the next state of the RM cannot be determined solely from an observed transition from the environment (\emph{e.g.,} \cite{toro2019learning,icarte2019searching,xu2020joint,xu2021active,furelos2021induction,furelos2020induction,rens2020learning}). Similarly, efforts have been directed toward leveraging the concept of probabilistic RMs (\emph{e.g.,} \cite{dohmen2022inferring,alam2022reinforcement,murphy2022probabilistic,lin2024efficient}) and stochastic RMs (\emph{e.g.,} \cite{corazza2022reinforcement,neider2022reinforcement}), with reward shaping \cite{ng1999policy} applied to RMs \cite{icarte2022reward,camacho2017non,jiang2021temporal}.

However, none of the mentioned results integrate physical dynamics into the formulation of reward structures and tasks for agents within their RM framework. Our work aligns with the increasing adoption of physical dynamics in the learning process through neural networks, as demonstrated in the concept of PINNs as in
\cite{cai2021physics,cuomo2022scientific,raissi2017inferring,raissi2017machine,pang2019fpinns,krishnapriyan2021characterizing,raissi2019physics} and the references therein. Here, we introduce additional information in the form of encoded dynamics related to the control tasks and reward structures, and propose pRM as symbolic RL rewarding machines, which are aware of those extra physical information. This enhancement is designed to give the pRM greater expressive power than existing notions of RMs in the literature, which can be extended over infinite regular languages. This increased expressiveness can be exploited to facilitate the learning of optimal policies, enabling the agent to maximize rewards more rapidly.

\section{Some Notable Remarks}
\begin{remark}
It should be noted that while the definition of pRMs in Definition \ref{rm} may appear syntactically similar to that of RMs in \cite{icarte2022reward}, there are fundamental differences in their operations. Unlike the finite-state machines described in \cite{icarte2022reward}, pRMs are not necessarily finite-state. Their transitions incorporate the ODE described in \eqref{prm_ode}, which captures task-related physical information---a concept absent in \cite{icarte2022reward}. Moreover, by definition, pRMs exhibit a hybrid configuration, a characteristic not present in \cite{icarte2022reward}. This hybrid structure allows pRMs to model non-Markovian reward structures, temporally extended behaviors, and time-constrained logics.
\end{remark}

\begin{remark}
    We note that pRMs are inspired by the notion of hybrid automata---a well-established formalism used to specify logical properties of hybrid dynamical systems, as discussed in works such as \cite{henzinger1996theory,lygeros2003dynamical}. Traditional hybrid automata typically provide qualitative judgments based on discrete transitions, indicating whether a system satisfies a given specification. However, they often rely on abrupt changes in the environment's state to drive these transitions. In contrast, pRMs go beyond such binary evaluations by assigning reward structures based on both discrete transitions and continuous physical cues derived from past experiences. A key advantage of pRMs is their ability to capture evolving progress signals (via the flow in \eqref{prm_ode}) even when the observable environment state remains unchanged.
\end{remark}

\section{Experimental Setups}
\begin{figure}[t!]
        \centering
       \includegraphics[width=0.65\linewidth]
{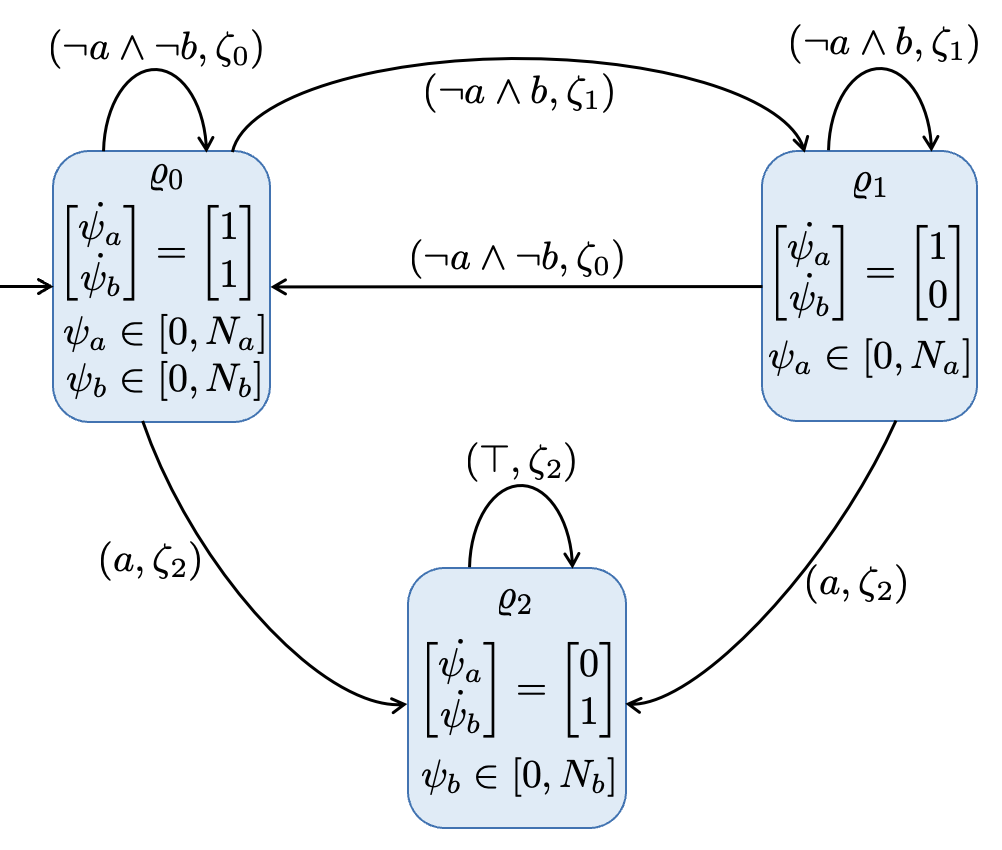}
\caption{pRM $\mathcal{A}_{R1}$.} 
        \label{fig_rm1}
\end{figure}

We first introduce other examples of pRM, which will be used in the upcoming experiments.
\begin{example}
    \label{eg_rm1}
    Consider the pRM $\mathcal{A}_{R1}$ depicted in Figure~\ref{fig_rm1}. The set of atomic propositions $\Delta=\{a,b\}$, the set of states $\Omega=\{\varrho_i~|~i\in[0;2]\}\times([0,N_a]\times[0,N_b])$, where $N_a,N_b\in\mathbb{N}_{\ge2}$. The terminal states set $\Omega_F=\{(\varrho_2,(\psi_a,\psi_b))~|~(\psi_a,\psi_b)\in[0,N_a]\times[0,N_b]\}$ and the initial state set $\Omega_0=\{(\varrho_0,(0,0))\}$. Each edge of the graph in Figure~\ref{fig_rm1} representing $\mathcal{A}_{R1}$ has a pair of the form $(\varphi,\zeta)\in 2^\Delta\times\mathbb{R}$. So, $\varphi\in2^\Delta$ is the propositional symbol that $\mathcal{A}_{R1}$ observes from the environment, which it utilizes to make a transition, and $\zeta\in\mathbb{R}$ is the reward assigned to the agent based on that transition. For any $i\in[0;2]$, the corresponding ODE, according to \eqref{prm_ode}, is specified at each node of the pRM depicted in Figure~\ref{fig_rm1} and we sample the flow at time-step intervals $\tau_{\mathcal{A}}=1$. As a result, one could easily obtain a discretized finite version of pRM $\mathcal{A}_{R1}$. We note that a labeling function assigns symbols from $\Delta$ to certain regions of the environment state set. Consequently, $\mathcal{A}_{R1}$ specifies the following tasks: the agent always visits the region labeled $b$ while avoiding $a$ and ensuring that no visit to $b$ lasts more than $N_b$ time steps \emph{i.e.,} $(\Delta b^{\le N_b})^\omega$ in an infinite word language form.
\end{example}
\begin{figure}[!ht]
        \centering
        \includegraphics[width=1.05\linewidth]{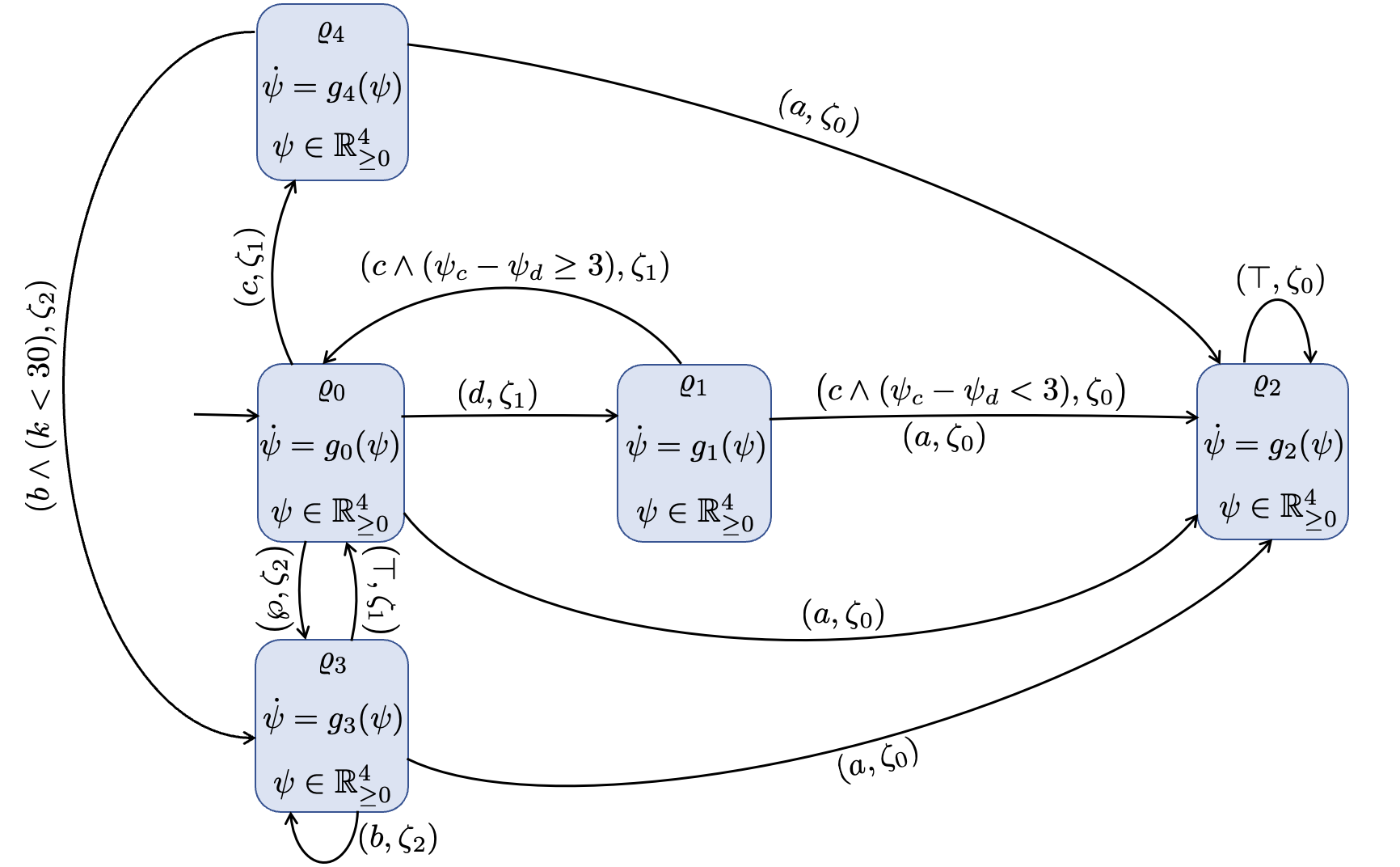}
\\[2.5ex]
    \caption{pRM $\mathcal{A}_{R3}$, where $\psi=[\psi_a;\psi_b;\psi_c;\psi_d]$, and $\wp$ represents $b\land(k-\psi_c\in[0,10])$.}
    \label{fig_rm3}
\end{figure}
\begin{example}
    \label{eg_rm3}
Consider Figure~\ref{fig_rm3}, which depicts a pRM $\mathcal{A}_{R3}$. The set of atomic propositions is $\Delta=\{a,b,c,d\}$, and let $L$ be the state labeling function. The set of states $\Omega=\{\varrho_i~|~i\in[0;4]\}\times[0,N_a]\times[0,N_b]\times[0,N_c]\times[0,N_d]$, where $N_a,N_b,N_c,N_d\in\mathbb{N}_{\ge2}$. The terminal states set is $\Omega_F=\{(\varrho_2,(\psi_a,\psi_b,\psi_c,\psi_d))\}\subset\Omega$, and the initial state is $\tilde\varrho_0=(\varrho_0,(0,0,0,0))$. Each edge in the graph representing $\mathcal{A}_{R3}$ in Figure~\ref{fig_rm3} corresponds to a pair consisting of the observed proposition from the environment (using the labeling function $L$) and the assigned reward. The associated ODEs \eqref{prm_ode}, which are sampled at time-step intervals $\tau_{\mathcal{A}}=1$ are given as follows: $g_1(\psi)=[1~\;1~\;1~\;0]^\top$, $g_2(\psi)=[0~\;1~\;1~\;1]^\top$, $g_3(\psi)=[1~\;0~\;1~\;1]^\top$ and $g_0(\psi)=g_4(\psi)=[1~\;1~\;0~\;1]^\top$.

Therefore, pRM $\mathcal{A}_{R3}$ specifies that the RL agent must always follow the logical task specification $t1\land(t2\lor t3\lor t4)$ where: $t1$ is `do not reach $a$ while aiming $b$'; $t2$ is `after reaching $d$, do not visit $c$ for the next $3$ time steps'; $t3$ is `visit $b$ within $10$ time steps of visiting $c$'; $t4$ is `within $N_b$ time steps, $b$ must be visited after visiting both $d$ and $c$.'
\end{example}

We now consider a finite case involving our motivating example and a $2$-tank model with pRMs $\mathcal{A}_{R1}$ and $\mathcal{A}_{R2}$, respectively. Algorithm \ref{alg1} is applied with the following hyper-parameters: discount factor $\lambda=0.9$, learning rate $\kappa=0.5$, and exploration value $\varepsilon=0.1$. For all $(x,\tilde\varrho,u)\in X\times \Omega\times U$, the $\tilde q(x,\tilde\varrho,u)$-values are optimistically initialized to $2$. We run $65$ independent trials and report the median, along with the $25th$ and $75th$ percentile of the normalized average reward per step.  

Furthermore, we examine environments modeled by $5$-room temperature control and $5$-road network models, which feature continuous and high-dimensional spaces. The agent tasks in this case is described by pRM $\mathcal{A}_{R3}$ in Figure~\ref{fig_rm3}. We apply Algorithm \ref{alg2} with the following hyper-parameters: discount factor $\lambda=0.99$ and a network learning rate $\iota=0.0001$. The neural network is a feed-forward model with $3$ hidden layers, each containing $1024$ ReLU units. The target network is updated every $300$ steps, and at each step, the Q-functions are updated using $128h$ batched experiences from a replay buffer of size $50000h$, where $h=\lvert\Omega\setminus\Omega_F\rvert$ when using pRMEs, and $h=1$ for ordinary DDPG. For each experiment, we conduct $5$ independent trials and report the median, along with the $25th$ and $75th$ percentile of the normalized average reward per step. The following methods are implemented in our experiments for discrete domains: Q-Learning (QL), Q-Learning with reward shaping (QL+RS), Q-Learning using counterfactual experiences as described in Algorithm \ref{alg1} (pRME), and its reward shaping variant (pRME+RS). For continuous domains, the methods include: DDPG, DDPG with reward shaping (DDPG+RS), the use of pRMEs from Algorithm \ref{alg2} (pRME), and its reward shaping version (pRME+RS). 

We conducted a PINN-enhanced version of the Grid World environment, as shown in Figure \ref{fig_abla}. The PINN was designed to model the transition probabilities of the discrete system, replacing black-box observation sampling for the pRM. States and actions were represented using one-hot encoding, and the network architecture comprised three fully connected layers: two hidden layers with ReLU activations and an output layer with a softmax activation. This ensured the output represented a valid probability distribution over the next states. The network took a concatenated vector of the one-hot encoded state and action as input and predicted the probabilities of transitioning to all possible next states. During training, the PINN minimized the difference between predicted and ground-truth transition probabilities using the Mean Squared Error (MSE) loss function. The optimization process employed the Adam optimizer for efficient gradient updates.

All implementations were executed on a $10$-core MacBook Pro ($3.2$ GHz, $64$GB RAM) with GPU acceleration using the MPS framework for Algorithm \ref{alg2}.
  
\paragraph{Grid world Environment.}
We conclude the grid world example described in Example \ref{mot_ex} by applying QL in this environment with the pRM illustrated in Figure~\ref{fig_rm2}. The associated parameters used in \eqref{rm2_ode} are $T_e=20$$^\circ C$, $T_0=98$$^\circ C$ and $\alpha=3.3\times10^{-4}$ per second. The corresponding reward values for pRM $\mathcal{A}_{R2}$ are $\zeta_0=\zeta_2=0$ and $\zeta_1=1$. Figure~\ref{fig_gridworld_rew} shows the performance of the respective rewards structure exploitation schemes in the pRM.

\paragraph{2-Tank Model.}
Consider the tank system \eqref{2tank} arranged in cascade, which is an adaptation of the tank model presented in \cite{ajeleye2024data}.
 \begin{align}
\label{2tank}
x_1(k+1)&=\big[\sqrt{\beta^2+x_1(k)+\tau u}-\beta\big]^2+0.01\varpi_1(k),\notag\\
    x_2(k+1)&=\big[\sqrt{\beta^2+x_2(k)+\tau \sqrt{x_{1}(k+1)}}-\beta\big]^2\notag\\
    &\;+0.01\varpi_2(k),
    \end{align}
where $\beta=0.5\tau$ and $\tau=10s$ is the sampling time, $\varpi_1$ and $\varpi_2$ are additive Gaussian noise of zero mean and variance $0.01$. The state $x_i(k)$ and $\sqrt{x_i(k)}$ represent the level of fluid and the outflow rate of the $i$-th tank, respectively, at time $k\in\mathbb{N}$. Here, the agent's task is specified by pRM $\mathcal{A}_{R1}$ in Figure~\ref{fig_rm1}. The rewards value $\zeta_0=\zeta_2=0$ and $\zeta_1=1$ with $N_a=N_b=10$. The set of states $X=[0,100]^2$ with the inflow rate $u$ taking values from the set of actions $\{0,1.5,4.5,7.5,9\}$. $L:X\rightarrow\{a,b\}$ is a labeling function defined as: $L(x)=b~\forall x\in[20,70]^2$ and $L(x)=a~\forall x\in [0,0.5]^2\cup[80,100]^2$. Figure~\ref{fig_2tank_rew} shows the performance of reward structure exploitation schemes in the pRM.

\paragraph{Room Temperature Model.}
Here, we consider a cyclic network of $5$-room temperature model, where each room is equipped with a heater. The model is borrowed from \cite{lavaei2020amytiss}, which shows the temperature evolution for each room as follows:
\begin{equation}\label{5rm_model}
    \begin{split}
        x_i(k+1)&=\theta_ix_i(k)+\varsigma T_hu_i+\xi \gamma_i+\beta T_e+0.01\varpi_i(k)\\
        &~\;\;\:\,\,i\in\{1,3\},\\
        x_i(k+1)&=\phi_ix_i(k)+\xi\gamma_i+\beta T_e+0.01\varpi_i(k),\\
        &~\;\;\:\,\,i\in\{2,4,5\},
    \end{split}
\end{equation}
where $\varpi_i$ is some additive noise, $\theta_i=(1-\beta-2\xi-\varsigma u_i)$, $\phi_i=(1-\beta-2\xi)$, and $\gamma_i=T_{x_{i+1}}+T_{x_{i-1}}$ such that $T_{x_{1}}=T_{x_{6}}$ and $T_{x_5}=T_{x_0}$. The heater and outside temperatures are represented by $T_h=50^\circ C$ and $T_e=-1^\circ C$, respectively. The parameters $\varsigma=0.05$, $\beta=0.022$ and $\xi=0.3$ are conduction factors between the $i$-th room and the heater, the external environment and the $(i+1)$-th room, respectively. The set of states $X=[15,25]^{5}$ and control actions $u_1$ and $u_3$ of the first and $3$rd room, takes values from set $[0,1]$.

The agent tasks are specified by $\mathcal{A}_{R3}$ in Figure~\ref{fig_rm3}, where $N_a=N_b=N_c=N_d=30$ and the rewards $\zeta_0,\zeta_1$ and $\zeta_2$ are, respectively, $0,0$ and $1$. The labeling function $L:[15,25]^{5}\rightarrow\{a,b,c,d\}$ is defined as follows: $L(x)=a~\forall x\in[15,18.5]^5\cup[21.5,25]^{5}$, $L(x)=b~\forall x\in [19.5,20.5]^{5}$, $L(x)=c~\forall x\in (18.5,19.5)^{5}$
        and $L(x)=d~\forall x\in (20.5,21.5)^{5}$.

Figure~\ref{fig_5rm_rew} shows the performances of our rewards structure exploitation schemes over the pRM.

\paragraph{Road Traffic Model.}
Our last experiment is a road traffic network that has $2$ entries and exits and is divided into $5$ cells of length $L=0.5km$ each. The model is borrowed from \cite{le2013mode}, and includes some stochasticity as additive zero mean noise with variance $0.7$. The entries are controlled by traffic lights, such that green light (\emph{i.e.,} $u_1$) allows vehicles to pass with a flow speed $v=100km/hr$, while red lights (\emph{i.e.,} $u_3$) halts them. Within a sampling time $\tau=6.48s$, we assume that $6$ and $8$ vehicles, respectively, pass the entries controlled by $u_1$ and $u_3$, while a quarter of vehicles that exit cells $1$ and $3$ goes through exit $1$ with a ratio denoted by $q$. The density $x_i$ of the traffic at the $i$-th cell of the road evolves as follows: 
\begin{equation}\label{5rd_net_rew}
    \begin{split}
        x_1(k+1)&=(1-T_v)x_1(k)+T_v\gamma_1(k)+6u_1+0.7\varpi_1(k)\\
        x_i(k+1)&=(1-T_v-q)x_i(k)+T_v\gamma_i(k)+0.7\varpi_i(k),\\
        &~\;\;\:\,\,i\in\{2,4\},\\
        x_3(k+1)&=(1-T_v)x_3(k)+T_v\gamma_3(k)+8u_3+0.7\varpi_3(k)\\
        x_5(k+1)&=(1-T_v)x_5(k)+T_v\gamma_5(k)+0.7\varpi_5(k)\\
    \end{split}
\end{equation}
where $T_v=\frac{\tau v}{L}$, $\gamma_i(k)=x_{i-1}(k)$ such that $x_{0}=x_{5}$. The set of states $X=[0,10]^{5}$ and control actions $(u_1,u_3)\in[0,1]^2$. The agent tasks are specified by $\mathcal{A}_{R3}$ in Figure~\ref{fig_rm3}, where $N_a=N_b=N_c=N_d=30$ and the rewards $\zeta_0=\zeta_1=0$ and $\zeta_2=1$. The labeling function $L:[0,10]^{5}\rightarrow\{a,b,c,d\}$ is defined as follows: $L(x)=a~\forall x\in(\mathbb{R}\setminus X)^{5}$, $L(x)=b~\forall x\in [1,8]^{5}$, $L(x)=c~\forall x\in (0,1)^{5}$ and $L(x)=d~\forall x\in (8,10)^{5}$.

Figure~\ref{fig_5rd_net_rew} shows the performance of our rewards structure exploitation schemes.


\end{document}